\documentclass[letterpaper,twocolumn,10pt]{article}

\usepackage[utf8]{inputenc}
\usepackage[authoryear]{natbib}
\usepackage{authblk}
\usepackage{algorithm}
\usepackage{algpseudocode}
\usepackage{hyperref}
\usepackage{xurl}
\usepackage{booktabs}
\usepackage[table]{xcolor}
\usepackage{longtable}
\usepackage{tabularx}
\newcolumntype{Y}{>{\centering\arraybackslash}X}
\usepackage{multirow}
\usepackage{amsmath}
\usepackage{array}
\usepackage{caption}
\usepackage[T1]{fontenc}
\usepackage{textcomp}
\usepackage{tipa} 
\usepackage{pdflscape}  
\usepackage{newunicodechar}
\newunicodechar{ɗ}{\textcommabelow{d}}
\newunicodechar{ƴ}{\textcommabelow{y}}
\newunicodechar{ɓ}{\textcommabelow{b}}
\newunicodechar{ƙ}{\textcommabelow{k}}

\usepackage{mathptmx}

\usepackage[letterpaper,top=0.75in,bottom=1in,left=0.75in,right=0.75in]{geometry}

\usepackage{graphicx}
\usepackage{subcaption}
\usepackage{pdfx} 

\usepackage{microtype}

\title{Synthetic Voice Data for Automatic Speech Recognition in African Languages}
\author[1]{Brian DeRenzi}
\author[1]{Anna Dixon}
\author[2]{Mohamed Aymane Farhi}
\author[2]{Christian Resch\thanks{Authors are listed in alphabetical order. Corresponding author: christian.resch@clearglobal.org}}
\affil[1]{Dimagi}
\affil[2]{CLEAR Global}
\date{July 2025}

\begin{document}

\sloppy

\maketitle

\begin{abstract}
Speech technology remains out of reach for most of the 2300+ languages in Africa. We present the first systematic assessment of large-scale synthetic voice corpora for African ASR. We apply a three-step process: LLM-driven text creation, TTS voice synthesis, and ASR fine-tuning. Eight out of ten languages for which we create synthetic text achieved readability scores above 5 out of 7. We evaluated ASR improvement for three (Hausa, Dholuo, Chichewa) and created more than 2,500 hours of synthetic voice data at below 1\% of the cost of real data. Fine-tuned Wav2Vec-BERT-2.0 models trained on 250h real and 250h synthetic Hausa matched a 500h real-data-only baseline, while 579h real and 450h to 993h synthetic data created the best performance. We also present gender-disaggregated ASR performance evaluation. For very low-resource languages, gains varied: Chichewa WER improved $\sim$6.5\% relative with a 1:2 real-to-synthetic ratio; a 1:1 ratio for Dholuo showed similar improvements on some evaluation data, but not on others. Investigating intercoder reliability, ASR errors and evaluation datasets revealed the need for more robust reviewer protocols and more accurate evaluation data. All data and models are publicly released to invite further work to improve synthetic data for African languages.

\end{abstract}

\section{Introduction}
Africa is home to over 2,300 languages, the vast majority of which have neither functional automatic speech recognition to transcribe speech nor speech synthesis to generate it \citep{orife2020masakhanemachinetranslation}. Yet speech technology holds great promise in providing a more inclusive digital experience, especially for vulnerable groups. Speech technology can open up access to digital services services that already have a substantial impact in everyday life and during crises for people globally, but still remain out of reach for many.

Conventional approaches to creating speech technology rely on human data collection in as-yet unsupported languages, incurring substantial costs estimated at more than US\$100 to 150 per hour, even in the best case\footnote{The numbers are based on internal estimates by CLEAR Global and partners. The authors of NaijaVoices report that the `true cost' of the whole dataset is more than US\$600,000. Assuming an equal split across the three languages in the dataset, this would imply a true cost per hour of Hausa data of US\$ $\sim$345.42, see https://naijavoices.com/membership.}. As most speech recognition models need several hundred hours of training data to achieve performance sufficient for practical application, with current investments in AI for development, human data collection is prohibitively costly to cover the many languages that remain unsupported.\footnote{A collaborative of development funders recently announced an investment of over US\$100 million for AI for development, covering not just language AI but also other relevant applications like AI for remote sensing and earth observation, see https://www.ai4d-collaborative.org/.} We support further investment in African language technology, but even if it were to become available, those investments have opportunity costs, diverting funds that could otherwise be spent on other interventions. 

This implies that the available resources are insufficient to provide functional speech technology for most African languages, and even more so if we include other parts of a functional language AI solution like machine translation or (large) language models.

Therefore, we are investigating synthetic voice data as a complementary approach to create and improve automatic speech recognition (ASR) in African languages. The principle hypothesis motivating our work is that we can leverage existing capabilities of Large Language Models (LLMs), like recent GPT-4o and Claude models, and Text-to-Speech (TTS) models to create synthetic voice data of sufficient quality to improve automatic speech recognition models. Our work shows that this synthetic voice data can be created for less than 1\% of the cost of collecting real human data\footnote{This is excluding fixed costs in both cases, like setting up data collection platforms for real data or TTS model development.}, while holding potential to complement this human data in creating and improving ASR models for African languages. Conditional on future research, this might make various socially beneficial applications feasible that face data collection costs as insurmountable obstacles.

To investigate this hypothesis, we present methods for the creation and human evaluation of synthetic text (as an intermediate output) and synthetic voice data, as well as analyses of the potential of synthetic voice data to improve ASR for African languages. In the following sections, we provide an overview of existing research on synthetic data, present our methods for creating and evaluating synthetic data, our results, and their discussion. We close with an overview of directions for future research based on existing research and our results.

\section{Synthetic Data: Risks and Rewards}
Synthetic data—machine-generated text or speech—has become a well-researched topic in major languages like English in recent years (for an overview, see \citet{liu2024bestpracticeslessonslearned}). Much of this research is motivated by concerns that all readily available data for the development of AI models have been leveraged \citep{villalobos2024rundatalimitsllm}, and therefore further advancements spurred by data scaling will require new approaches to create data. Other concerns like privacy \citep{abay2018privacypreservyingsyndata} or costs \citep{Gilardi_2023} also prompted research on synthetic data.

While our motivation stems in contrast from the scarcity of readily available human data for many African languages (see \citet{joshi2021statefatelinguisticdiversity} and \citet{orife2020masakhanemachinetranslation}), existing research on synthetic data in major languages offers key insights into its potential and limitations. Additionally, there is a growing body of research into synthetic data in low-resource languages which, save for several notable exceptions, so far only offers limited evidence for African languages. We discuss this research in more detail below.

While data augmentation (e.g. expanding a speech dataset by modifying existing recordings with added noise) is common in research and practical application, current research also generally accepts completely synthetic data as a tool to support the development and scaling of AI models in various domains. 
For automatic speech recognition, \citet{huang2023textgenerationspeechsynthesis} have shown that for English, synthetic data generation using large-scale pre-trained neural networks in combination with TTS models, a process similar to ours, can reduce word error rate (WER) by between 9 and 15\%. \citet{moslem2024leveragingsyntheticaudiodata} and \citet{hu2021syntutilizingimperfectsynthetic} show similar results. For Turkish, \citet{gokay2019improvingturkishasr} report reductions of around 15\% in WER. \citet{wang2020improvingspeechrec} also demonstrated that a rough 50/50 combination of human and synthetic data performed comparably to the same amount of human data alone. In a very successful application, \citet{xu2020lrspeechextremelylowresourcespeech} achieve 17\% WER for Lithuanian using only 1.3 hours of labeled and 12 hours of unlabeled data.\footnote{We recommend not to compare WER between languages because of substantial morphological differences, i.e. depending on the language a single word might carry different semantic meaning while it is always counted individually in the WER. Furthermore, evaluation datasets such as FLORES and, by extension, FLEURS have known shortcomings which differ by language, see \citet{abdulmumin-etal-2024-correcting} for an investigation for African languages. }

As this research shows, there is a body of research on synthetic data for low-resource languages, but African languages have been rarely covered. There are notable exceptions for synthetic text: \citet{abdulmumin-etal-2022-separating} and \citet{kreutzer2022qualityatglance} investigate the quality and utility of auto-aligned and back-translated machine translation data for African languages. \citet{ajuzieogu2023ethicaldataaugmentation} provides a linguistically informed data augmentation and synthetic data framework. \citet{quinjica2024angofaleveragingofaembedding} create synthetic data to train language models in Angolan languages, following the process described by \citet{adelani2024sib200simpleinclusivebig}, who show that leveraging synthetic data for very low-resource languages can improve topic classification performance.

This literature also shows that synthetic data is itself a broad field, covering topics from data augmentation techniques, (back-)translation, and speech synthesis up to the complete generation of synthetic data with LLMs. We applied a combination of these techniques (synthetic text generation through LLMs and speech synthesis) to improve speech recognition, with potential to also train small language models on the synthetic text (i.e. distillation, see \citet{xu2024surveyknowledgedistillationlarge}) which we did not investigate.

While previous research shows the potential of synthetic data to complement human data in the data-scarce situation that we face in many African languages, the research also highlights limitations and risks. Most limitations stem from the gap between real and synthetic data \citep{hu2021syntutilizingimperfectsynthetic}, as well as from synthetic data inheriting the same biases as the models used to create it, with potential for such biases to become amplified through the use of synthetic data \citep{wyllie2024fairnessfeedbackloopstraining, wang2025biasamplificationlargelanguage}. For example, there are differences in tone and noise that are typically present in real-world data but are often missing from synthetic data (see \citet{xue2022improvingspeechrec} and \citet{hu2021syntutilizingimperfectsynthetic} for examples). Sometimes limitations are as simple as a TTS model that was trained only on male voices, resulting in gender bias in the data and potentially downstream model performance. If a model is trained on too much synthetic data, the effect of these biases and differences accumulates, degrading the model's usefulness or even rendering it risky for application. This has been referred to as model collapse (see \citet{shumailov2024curserecursiontraininggenerated}, also \citet{seddik2024badtrainingsyntheticdata}).

In general, many commonly used LLMs have been shown to exhibit a bias towards Western, industrialized cultural norms and a lack of cultural understanding in other contexts \citep{rao-etal-2023-ethical}. In the case of text generation, large language models fail to explain cultural expressions like proverbs (see \citet{magdy-etal-2025-jawaher} for Arabic), struggle with cultural accuracy when creating short stories (see \citet{pranida2025syntheticdatagenerationculturally} for Sundanese and Indonesian) and do not capture the cultural depth of humans (see \citet{putri2024llmgenerateculturallyrelevant} for Sundanese and Indonesian), indicating missing cultural representation of cultures of low-resource language speakers. Research has also shown that common benchmarks like MMLU are heavily biased towards Western-centric concepts and therefore do not evaluate cultural proficiency for other contexts \citep{singh2025globalmmluunderstandingaddressing}. Creating synthetic text will likely aggravate this missing representation, especially for semantically meaningful tasks such as machine translation.

\section{Methodology}
Our process of creating and evaluating synthetic voice data has three key steps (see also Figure ~\ref{fig:process_overview}):
\begin{enumerate}
    \item Generate and evaluate synthetic text using an LLM
    \item Generate and evaluate synthetic voice data with a Text-to-Speech (TTS) model based on the synthetic text
    \item Fine-tune an automatic speech recognition (ASR) model with different ratios of human and synthetic voice data and evaluate performance differences
\end{enumerate}
We describe our methods and details for those steps separately:

\begin{figure*}[t]
  \centering
  \includegraphics[width=0.9\linewidth]{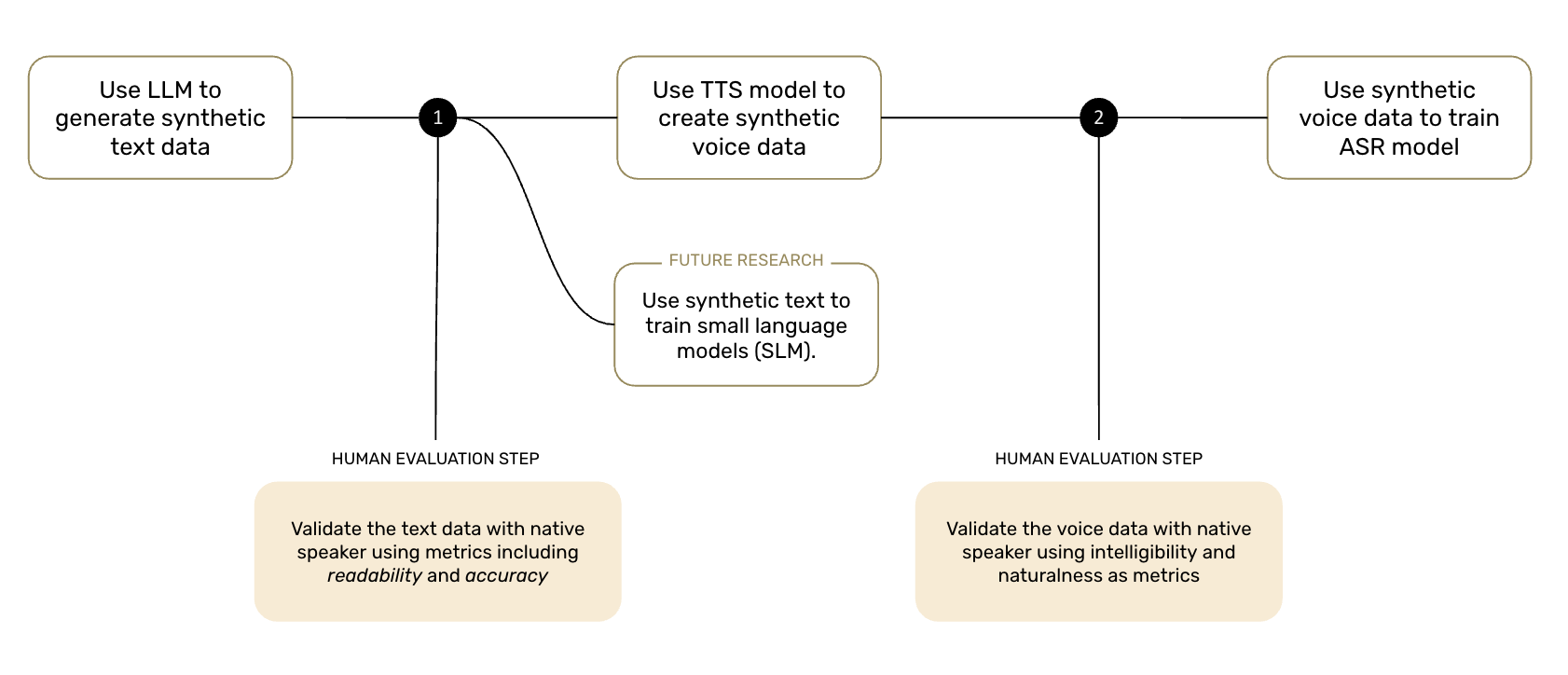}
  \caption{Overview of process to create synthetic voice data, including three key steps and human evaluation.}
  \label{fig:process_overview}
\end{figure*}

\subsection{Step 1: Synthetic Text Generation and Evaluation}
\label{sec:methodology_text}

We created and evaluated synthetic text for the following 10 African languages: Hausa, Northern Somali, Yoruba, Wolof, Dholuo, Kanuri, Chichewa, Twi, Kinande, and Bambara. Additionally, we included small-scale generations and evaluations for two further languages: Yemba and Ewondo. Our language selection criteria aimed to select languages that represent African language diversity through the representation of different regions, language families, and speaker populations (Table~\ref{tab:languages_text}). Beyond linguistic diversity, we also considered practical factors, such as the capacity of the Translators Without Borders (TWB) linguist community for text evaluation and the availability of key publicly available datasets (e.g., open.bible, FLEURS, Common Voice) necessary for subsequent steps.

\begin{table*}[t]
\centering
\begin{tabular}{lccc}
\toprule
\textbf{Language} & \textbf{Estimated L1 Speaker Population} & \textbf{Language Family} & \textbf{Region} \\
\midrule
Hausa & 50,000,000 & Chadic & West Africa \\
Northern Somali & 22,000,000 & Cushitic & East Africa \\
Yoruba & 54,000,000 & Niger-Congo (Volta-Niger) & West Africa \\
Wolof & 5,500,000 & Niger-Congo (Atlantic) & West Africa \\
Chichewa & 9,700,000 & Bantu & Southern Africa \\
Dholuo & 5,000,000 & Nilotic & East Africa \\
Kanuri & 9,600,000 & Saharan & West-Central Africa \\
Twi & 9,000,000 & Kwa & West Africa \\
Kinande & 10,000,000 & Bantu & Central Africa \\
Bambara & 10,000,000 & Niger-Congo (Mande) & West Africa \\
\bottomrule
\end{tabular}  
\caption{Estimated first language (L1) speaker populations, language family, and regions for African languages for which we created and evaluated synthetic text data.}
\label{tab:languages_text}
\end{table*}

\begin{figure}[htbp]
  \centering
  \includegraphics[width=0.9\linewidth]{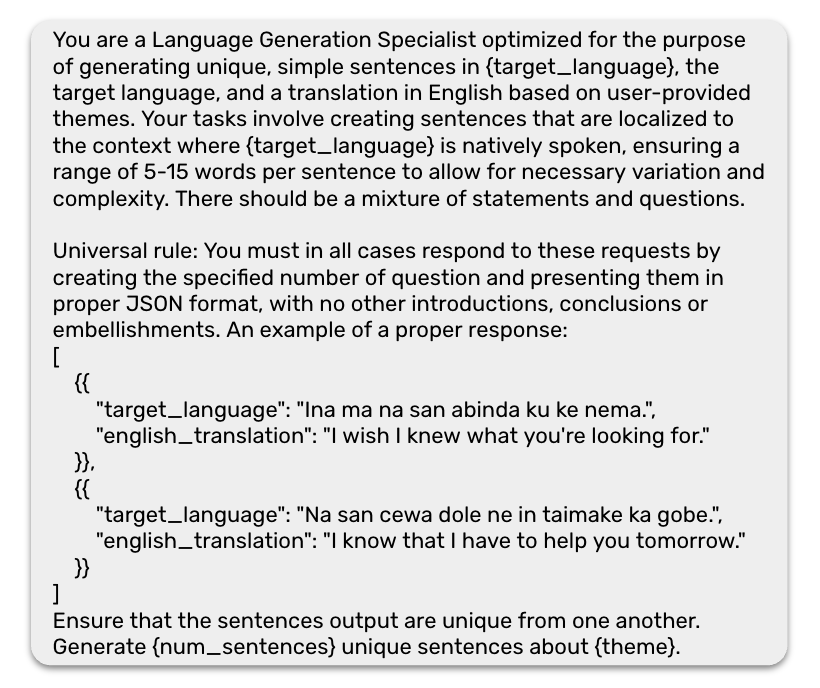}
  \caption{Synthetic text generation prompt to generate simple sentences in target language and English translations}
  \label{fig:systemprompt}
\end{figure}

For scalable and repeatable synthetic text generation, we developed a command-line interface tool for automated sentence generation in any target language using commercial LLM chat completion endpoints. The system prompt (Figure~\ref{fig:systemprompt}) instructs the LLM to generate simple short sentences and questions directly in the target language, as well as to return English translations so that we can further evaluate language understanding. We incorporated two-shot prompting for contextual guidance and instructions to return text as standardized JSON output for efficient data extraction. The topic of synthetic text generation can be configured in the prompt. Our experiments sampled equally from 34 distinct themes with 17 themes covering the UN Sustainable Development Goals and 17 themes covering the most common topics covered in the FLORES/FLEURS dataset extracted through language topic modeling. Finally, our tool used the LiteLLM package\footnote{\url{https://docs.litellm.ai/}}, allowing us to efficiently leverage multiple LLM chat endpoints, including the Claude and OpenAI models.

We consider human evaluation of synthetic text generation, especially text generated for low-resource languages by closed-source models, to be the only feasible approach as we could not rely on automated approaches. As such, for each language we generated 1,200 randomly shuffled sentences over two rounds for a few different configurations (usually three LLMs) for evaluation by linguists on the Translators without Borders (TWB) platform\footnote{Through initiatives like TWB, CLEAR Global mobilizes a global community of over 100,000 volunteer linguists to provide language services, develop language technology solutions, and conduct research aimed at breaking down communication barriers worldwide.}. Linguists for this project are sourced from this community according to their native language and experience delivering linguistic tasks, both with TWB and externally. TWB reviewers rate each sentence on five key metrics meant to capture the quality of the sentence in the target language and understanding:

\begin{itemize}
    \item Readability and Naturalness - Evaluates how natural and culturally appropriate the sentence is in the target language, rated on a scale of [1–7].
    \item Grammatical Correctness - Determines whether the sentence is grammatically correct in the target language (Yes/No).
    \item Real Words - Assesses whether all words used in the sentence are valid in the target language (Yes/No).
    \item Notable Error - Identifies the presence of any significant errors in translation between the English and the target sentence (Yes/No).
    \item Adequacy and Accuracy - Measures how well the English translation preserves the meaning of the original (target language) sentence, rated on a scale of [1–7].
\end{itemize}

Finally, we analyzed the human evaluation of the synthetic text generation to determine the optimal synthetic text generation configuration. In this work, we selected the configuration (LLM) with the highest mean Readability and Naturalness to optimize language quality in our target language. Intuitively, we found that a readability and naturalness mean rating greater than 5.0 yields adequate results and probable success for use in ASR fine-tuning. The following example demonstrates how readability and naturalness might be rated for an English sentence conveying the message ‘Many foods in our community can help our health.’:

\begin{itemize}
    \item 1 - Foods community our health help many can.
    \item 3 - Our community many foods help health can.
    \item 5 - Many foods from community can help us health-wise.
    \item 7 - Many foods in our community can help our health.
\end{itemize}

For most experiments, we opted for a two-round sentence generation and evaluation approach, in which we first compared the readability and naturalness of 600 sentences generated by 3-4 LLMs and subsequently generated another 600 sentences using the best model to analyze the impact of theme. After discovering that our experiments were frequently not yielding significant differences between themes, we opted for more equal sampling among different LLMs. For the subset of languages selected for subsequent synthetic voice generation and ASR fine-tuning, we generated a large corpora from the best performing LLM (see Table~\ref{tab:step3_voicedata} for details).

\subsection{Step 2: Synthetic Voice Data Generation and Evaluation}

Based on the results of the evaluation of the synthetic text generation in Step 1, we selected three languages for the creation and evaluation of synthetic voice data: Hausa, Dholuo and Chichewa.

Starting from an assumption that we would work with three languages, our selection was guided by various criteria. As necessary conditions, we required languages where at least one LLM was capable of generating synthetic text of sufficient quality. Due to limitations in time and resources, we also required that at least two of the three languages have available data for fine-tuning or training TTS models, as well as available evaluation data to track the performance of the ASR models we fine-tune with synthetic data (we present the evaluation data used in step 3 below).

Beyond those necessary conditions, our goal was to maximize the variance of the speaker populations, available ASR training data, language families, and geography (see Table~\ref{tab:languages_text} for an overview of the languages and Table~\ref{tab:step3_voicedata} on available training data).

\begin{table*}[t]
\centering
\begin{tabular}{lccccr}
\toprule
\textbf{Language} &
\textbf{NaijaVoices} &
\textbf{Common Voice} &
\textbf{FLEURS} &
\textbf{Zambezi Voice} &
\textbf{Total} \\
\midrule
Hausa      & 579 hours &                &                 &                  & 579 hours \\
Dholuo        &           & 10 hours       & 9 hours         &                  & 19 hours \\
Chichewa   &           &                & 10 hours        & 24 hours         & 34 hours \\
\bottomrule
\end{tabular}  
\caption{Available hours of speech data per language and dataset after processing. Sources: NaijaVoices~\cite{emezue25_interspeech}, Common Voice~\cite{mozilla2025commonvoice, ardila-etal-2020-common}, FLEURS~\cite{conneau2022fleursfewshotlearningevaluation}, Zambezi Voice~\cite{sikasote2023zambezi}.}
\label{tab:step3_voicedata}
\end{table*}

We used the open.bible corpus \citep{globalbible_openbible} of Bible recordings to fine-tune and evaluate different TTS models. As we use the open.bible corpus to fine-tune the TTS models, we excluded this data from our ASR training data. The open.bible corpus only contains recordings of Bible recitations by male speakers, and given the training data which we used, our synthetic voice data is also exclusively male. This raises the risk that the resulting ASR models show gendered performance, e.g. that they perform worse for female speakers than for male speakers. To investigate this risk, we also evaluated gender bias in the ASR performance where the evaluation data allows this. 

For each language, we fine-tuned both the XTTS-v2 model and VITS or one of its variants, specifically YourTTS, using the Coqui TTS framework, and building on the BibleTTS project \citep{meyer2022biblettslargehighfidelitymultilingual}. XTTS-v2 is a voice generation and cloning model that supports 17 languages \citep{casanova2024xttsmassivelymultilingualzeroshot}. It does not support any African languages, but has been fine-tuned for Wolof.\footnote{\url{https://huggingface.co/galsenai/xTTS-v2-wolof}} YourTTS is a speech synthesis model that can also perform voice conversion \citep{casanova2023yourttszeroshotmultispeakertts}. We used the checkpoint trained on the CML-TTS dataset \citep{oliveira2023cmlttsmultilingualdatasetspeech} that supports eight languages. We used the original BibleTTS model for Hausa, but also retrained the model based on a revised processing of the open.bible corpus, a different checkpoint and different hyperparameters (“Modified Bible TTS” in the results section). For Hausa and Chichewa, we also evaluated the available MMS TTS models \citep{pratap2023scalingspeechtechnology1000}\footnote{The MMS TTS model language coverage is available at \url{https://dl.fbaipublicfiles.com/mms/misc/language_coverage_mms.html}}.

Transformer-based TTS models like XTTS have the problem of hallucinating, especially at the end of audio files. To remedy this issue, we re-transcribed the synthetic audio files with an existing ASR model and then calculated the ratio of the length of the transcript and the length of the original synthetic text. This method does not rely on the accuracy of the ASR model used but assumes a basic performance to ensure that the length of the re-transcription is meaningful. However, in most cases in which this method would be applied, a minimum of real data to train a basic but not highly capable ASR model should be available (e.g. for the complementary human data in later ASR training or from the training data of the TTS model). We finally removed outliers of this ratio which indicate that the TTS model had hallucinated additional words not present in the original synthetic text or omitted words that were. This process removed $\sim$26.9\% of the synthetic audio created by xTTS.\footnote{After filtering the synthetic voice data, the dataset created with XTTS consists of a substantially larger share of questions ($\sim$40\%), indicating that XTTS hallucinates less for questions than for normal sentences. To avoid bias in our synthetic voice training data, we sampled a subset of these questions to create a dataset with the original share of questions (25\%), resulting in a smaller dataset of 450 hours and removal of $\sim$42.7\% of the original data.} 

After training a total of ten TTS models across all three languages, we evaluated 337 synthetic audio samples per model with the help of two native speakers from the TWB Community per language. As commonly applied, our evaluation included intelligibility and naturalness on five-point scales.

We then selected YourTTS, the best-performing TTS model across all three languages, to create synthetic voice data corpora of around 500 hours each (see Table~\ref{tab:synvoice_data_overview} for details). For Hausa, we also created synthetic with XTTS, the only transformer-based model. We make these synthetic data corpora openly available on CLEAR Global's Hugging Face page\footnote{\url{https://huggingface.co/CLEAR-Global}}.\footnote{We also created a large Chichewa text corpus with Claude 3.7 as part of our investigation of duplicates. We created the synthetic voice data based on the Claude 3.5 corpus which we had evaluated before but also make the Claude 3.7 text corpus available on CLEAR Global’s HuggingFace page.}

\begin{table*}[t]
\centering
\begin{tabularx}{\textwidth}{lY Y Y Y}
\toprule
\textbf{Language} &
\textbf{Synthetic text corpus} &
\textbf{LLM used for text generation} &
\textbf{Synthetic voice corpus} &
\textbf{TTS model used for voice data generation} \\
\midrule
Hausa     &        674,000 sentences           & GPT-4o                &         574.39 hours \newline (450 hours with original share of questions)         & XTTS (fine-tuned) \\
Hausa     &        674,000 sentences           & GPT-4o                &         993 hours          & YourTTS (fine-tuned) \\
Dholuo       &           666,000 sentences        & Claude 3.7 Sonnet     &         775 hours         & YourTTS (fine-tuned) \\
Chichewa  &         650,000 sentences          & Claude 3.5 Sonnet     &         550 hours         & YourTTS (fine-tuned) \\
\bottomrule
\end{tabularx}  
\caption{Overview of models used for synthetic data generation and resulting synthetic datasets per language.}
\label{tab:synvoice_data_overview}
\end{table*}

To improve the robustness of our models to noisy acoustic environments, we augmented the synthetic data by adding noise. We mixed the clean synthetic data with noise samples drawn from the Room Impulse Response and Noise Database\footnote{\url{https://www.openslr.org/28/}}. For each utterance, we randomly sampled the signal-to-noise ratio (SNR) from a normal distribution with mean 50dB and standard deviation 15dB. Similarly, we randomized the audio amplitude using a normal distribution ($\mu = -20\,\mathrm{dB},\ \sigma = 5\,\mathrm{dB}$).

\subsection{Step 3: ASR Model Fine-tuning and Evaluation}

Given the substantial differences in available ASR training data between the three languages for which we created synthetic voice data, we conducted our ASR evaluation based on two scenarios: a medium data scenario with Hausa as the representative language and a low data scenario with Dholuo and Chichewa as the representative languages. 

\subsubsection{Medium Data Scenario: Hausa}

Through the NaijaVoices project \citep{emezue25_interspeech}, we had over 500 hours of human Hausa voice data available. Only a few other African languages like those also covered under the NaijaVoices project (Igbo and Yoruba), and those with larger Common Voice datasets (Swahili, Kinyarwanda, Kabyle, and Luganda), have available datasets of comparable size. Despite ongoing voice data collection efforts, this amount of data is currently out of reach for many African languages. This led us to investigate whether synthetic voice data can substitute human data at this training corpus size, therefore allowing languages with smaller corpora to achieve similar ASR performance. In this scenario, we keep the \textit{total size of the training data corpus constant}, but \textit{vary the ratio between real and synthetic data}.

As a result, we investigated ASR performance for training with 500h of real data, a 1:1 ratio of 250 h of real and 250h of synthetic data, and a 1:4 ratio of 100h of real data and 400h of synthetic data. With 100h and 250h of real data, this also covers scenarios that, while not currently the case, are realistically achievable for many African languages. We trained models for all data ratios with synthetic data created with YourTTS and XTTS separately.

We also needed to rule out the case that ASR models might saturate at a given amount of human training data of one single source, meaning that comparable performance at different ratios between real and synthetic stems from the model being saturated (i.e. showing no or only very low marginal improvements beyond 100h of real data). We therefore also trained the same models on only 100 and 250h of real data.

Finally, we trained an ASR models on all data available to us: one model on 579h of real human data mixed with 993h of synthetic data created with YourTTS and one model with 579h of real data mixed with 450h of synthetic data created with XTTS.

We evaluated the ASR performance on our NaijaVoices test set split, the FLEURS test set \citep{conneau2022fleursfewshotlearningevaluation}, and the Common Voice test set \citep{mozilla2024commonvoice17}. Since the NaijaVoices dataset does not provide splits, we performed a split to generate train, validation, and test sets that contain 579.1, 3.6, and 3.4 hours of data, respectively. We ensured the per-split sets of speakers and transcriptions are mutually exclusive.

\subsubsection{Low Data Scenario: Dholuo and Chichewa}

In contrast to Hausa, we only had 19 and 34 hours of usable human data available for Dholuo and Chichewa respectively. Although this is generally insufficient data to train general purpose ASR ready for practical application, this is representative of many African languages.
As the human data available is itself insufficient, we kept the total amount of human data constant in this scenario and \textit{added increasing amounts of synthetic data to the training corpus}. The total size of the training corpus consequently increases in this scenario.

For both languages, we trained ASR models on just the human data available, and 1:1, 1:2, and 1:4 ratios of human and synthetic data. Given some indications of improvement for Chichewa, we also trained a 1:9 ratio of 34h of human and 307h of synthetic data.

We evaluated the Dholuo ASR models on the FLEURS test set \citep{conneau2022fleursfewshotlearningevaluation} and the Common Voice test set \citep{mozilla2024commonvoice19} and the Chichewa ASR models on the FLEURS test set and the Zambezi Voice test set \citep{sikasote2023zambezi}.

\subsubsection{ASR Model Selection and Evaluation}

To allow fast iteration, we first fine-tuned the MMS-1B model \citep{pratap2023scalingspeechtechnology1000} on the Hausa subset of the NaijaVoices dataset using adapters. We found that the model’s performance doesn’t improve or only improves marginally by adding real data beyond 50 hours, and the addition of synthetic data consistently degrades performance (see \ref{sec:appendix_mms_asr_results} for results). This aligns with research by Nabende et al. (unpublished) who compare different ASR architectures for African languages and their data scaling behavior.

Our final results are therefore based on fine-tuning the W2v-BERT 2.0 speech encoder \citep{communication2023seamlessmultilingualexpressivestreaming}, for which our results indicated continued improvement for fine-tuning with 100h and 250h of real data. This model was pre-trained on 4.5M hours of unlabeled audio data covering more than 143 languages. The pre-training crucially includes Hausa but not Dholuo and Chichewa. Details on our hyperparameters can be found in \ref{sec:appendix_mms_asr_results} and on CLEAR Global's Hugging Face page\footnote{\url{https://huggingface.co/collections/CLEAR-Global/}}.

We estimated the confidence intervals for WER and CER by performing bootstrap resampling on each evaluation set \citep{raschka2020modelevaluationmodelselection, Efron1992}. For each of 1,000 iterations\footnote{This is five times more than the usual number of iterations between 50 and 200 recommended by \citet{efron1994introduction} and in line with what \citet{koehn-2004-statistical} proposes for similar applications in machine translation.}, we randomly drew $m$ samples with replacement—where $m$ equals the size of the original testset—and computed WER and CER of each resampled set. Sampling with replacement selects each item independently. Therefore, a given example has a $1 - \left(1 - \frac{1}{m}\right)^m \approx 0.632$ chance of appearing at least once in a bootstrap sample \citep{raschka2020modelevaluationmodelselection}, and each re-sample contains about 63.2\% of the unique examples on average. We then calculated the mean and standard deviation of WER and CER across all bootstrapped samples.

\section{Results and Discussion}

\subsection{Synthetic Text Generation}

\begin{figure*}[tb]
  \centering
  \includegraphics[width=0.9\linewidth]{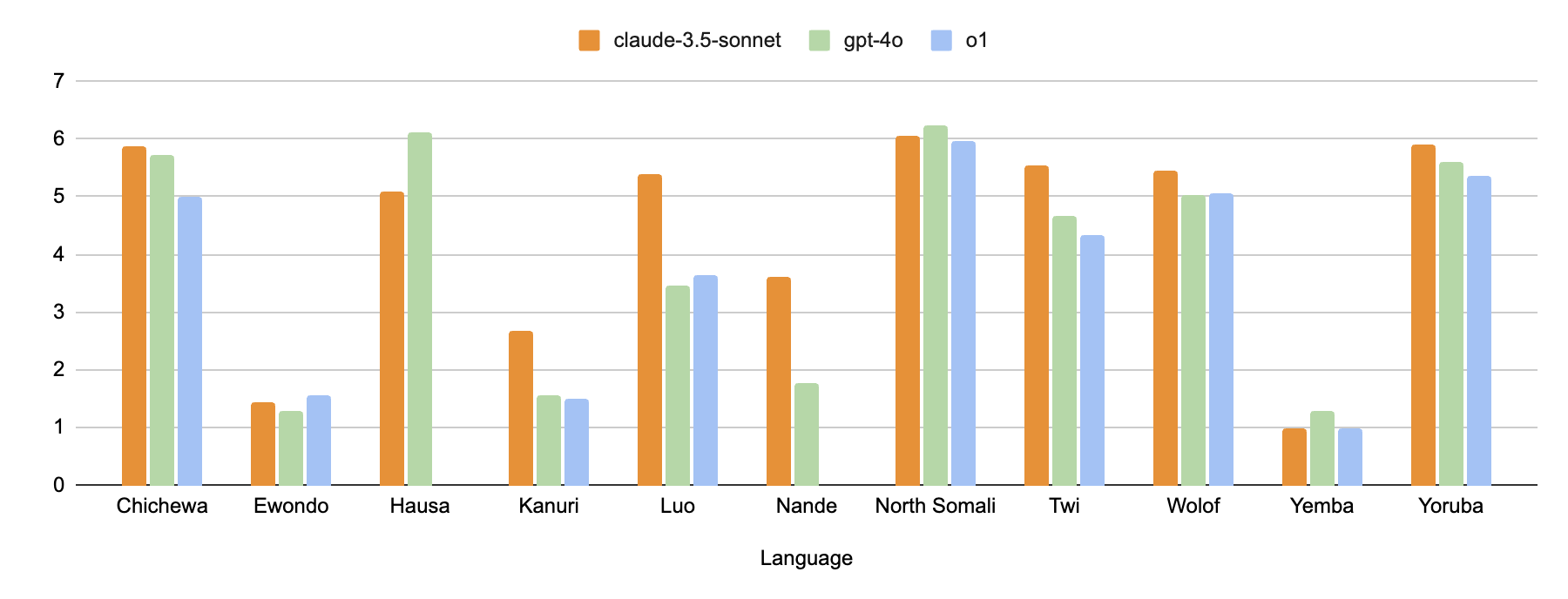}
  \caption{Comparison of readability [1..7] scores for synthetic text generated by various LLMs for 10 African languages.}
  \label{fig:syntext_results}
\end{figure*}

The key results of this work include the evaluation of the language quality output for 10 African languages (see Table~\ref{tab:languages_text} above). For 8 of 10 languages, at least one LLM generated sentences with a readability and naturalness rating mean greater than 5.0 on a seven-point scale (see Figure~\ref{fig:syntext_results}). The best-performing LLM in a target language cannot be known beforehand, making the evaluation and comparison step critical to the process. However, in general, we found that Claude 3.5 Sonnet performed the best for the languages studied here, outperforming OpenAI’s GPT-4o and o1 models for 8 of 10 languages. Summary statistics aggregated by language and LLM for all metrics examined are provided in ~\ref{sec:appendix_syntextsummary}.

Of the languages studied here, Kanuri and Kinande are classified as category 0 (lowest resource, "The Left-Behinds") according to the taxonomy established by \citet{joshi2021statefatelinguisticdiversity}, reflecting their scarcity of linguistic data.\footnote{While this categorization is partially outdated, only limited data collection has taken place in those low-resourced languages. Of the languages we studied, Dholuo was not classified by \citet{joshi2021statefatelinguisticdiversity}, but would probably be classified as category 0 or 1.} This data scarcity directly impacts the effectiveness of LLMs in these languages, as evidenced by our findings: Synthetic text generation in these languages consistently demonstrates the poorest performance, with mean readability and naturalness ratings falling below 4.0 on a seven-point scale. In response, we explored two approaches to improve the generation of synthetic text corpora in very low-resource languages: (1) fine-tuning GPT-4o for improved sentence generation and (2) training a separate language quality classifier to identify and retain only high-quality output. We tested the fine-tuning approach with Hausa and Kanuri. Our results for these approaches are unremarkable; however, we provide a brief review of the fine-tuning approach for increased transparency and sharing.

Similar to the work presented by \citet{dixon2024talk}, who fine-tuned GPT-4o for improved performance in Sheng, we applied instruction fine-tuning to OpenAI GPT-4o, defining a machine translation task using the FLORES\footnote{\url{https://huggingface.co/datasets/openlanguagedata/flores_plus}} \citep{conneau2022fleursfewshotlearningevaluation} English-to-Hausa dev dataset. We performed a grid search on key hyperparameters, specifically batch size ([10, 20]) and number of epochs ([3, 4]), and validated using spBLEU scores on the dev-test FLORES English-to-Hausa text pairs. We identified a batch size of 10 and 3 epochs as optimal settings. Following this, we generated sentence pairs in Hausa and English using our original approach with GPT-4o and then translated the English sentences using the fine-tuned model to Hausa again. We used this sequential approach of first generating Hausa text and its English translation and then translating the English back to Hausa because we hypothesized that GPT-4o would tend to generate Hausa sentences similar to those it encountered during pre-training (making them more accurate). Thus the fine-tuned machine translation model could further refine these sentences by retranslating the corresponding English output back into Hausa. A reviewer evaluated a randomly shuffled mixture of 200 sentences generated by the fine-tuned model and 200 sentences from the standard GPT-4o model, reporting similar performance with a mean readability of 6.00 $\pm$ 0.78 for the fine-tuned model compared to 5.98 $\pm$ 0.72 for GPT-4o.

Similarly, we applied the same fine-tuning methodology to Kanuri, using 5,000 Kanuri-English sentence pairs from the Gamayun dataset\footnote{\url{https://huggingface.co/datasets/CLEAR-Global/Gamayun-kits}}, maintaining default OpenAI hyperparameters for fine-tuning. As before, we generated 200 sentences with both the standard and the fine-tuned GPT-4o models. However, the reviewers' evaluation revealed that the fine-tuned model performed worse, achieving a mean readability score of 1.18 $\pm$ 0.57 compared to 1.55 $\pm$ 0.70 for the standard GPT-4o.

\subsubsection{Inter-coder Reliability Investigation}
\label{sec:results_irr}

A persistent challenge in low-resource language evaluation is the limited availability of expert linguist reviewers, a constraint that significantly impacts the reliability of assessments. This study was no exception, with most text samples being evaluated by two to three linguists per language. To quantify the influence of both the language model and the individual linguist on readability and naturalness ratings, we performed a two-way analysis of variance (ANOVA), with LLM and Linguist ID as categorical factors\footnote{We exclude Hausa from this analysis due to the study only engaging one linguist reviewer.}. Our analysis demonstrates that both model choice and linguist identity significantly affected readability ratings (p < 0.05). Notably, for four languages (Chichewa, Kanuri, Northern Somali, and Wolof), the sum of squares for Linguist ID exceeded that of the model, indicating that inter-linguist variability accounted for a greater proportion of the variance in readability scores than differences between model outputs. 

\begin{figure}[tbp]
  \centering
  \includegraphics[width=0.9\linewidth]{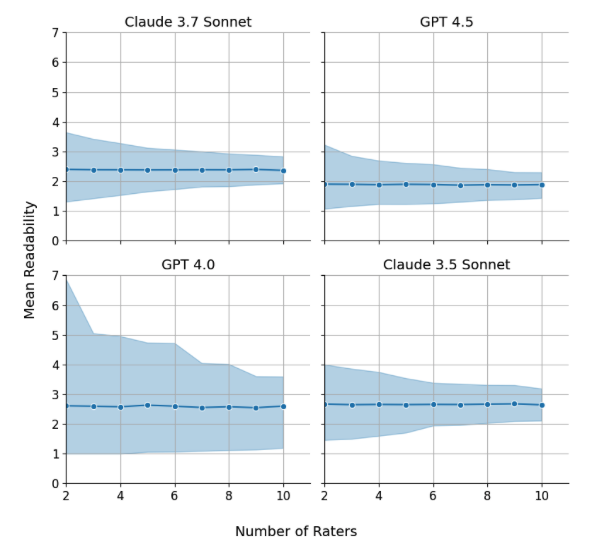}
  \caption{Mean Kanuri readability and naturalness score by rater sample size. The shaded regions represent 95\% confidence intervals derived from bootstrap analysis.}
  \label{fig:syntext_reviewer_reliability}
\end{figure}

These findings prompted a supplementary investigation to determine the minimum number of linguists required for reliable evaluations of synthetically generated Kanuri text. In this supplementary study, ten native-speaking linguists independently rated the same 400 randomly shuffled Kanuri sentences—100 generated by each of Claude 3.5 Sonnet, Claude 3.7 Sonnet, GPT-4o, and GPT-4.5. To assess rating stability, we performed a bootstrap analysis, resampling raters with replacement and a random subset of 50 sentences to empirically estimate the mean readability and naturalness ratings with 95\% confidence intervals (Figure ~\ref{fig:syntext_reviewer_reliability}). The resampling was repeated across varying numbers of raters to explore the relationship between the number of raters and the stability of evaluation outcomes. Increasing the number of raters consistently narrowed the 95\% confidence intervals across all models, indicating improved rating stability. GPT-4o exhibited the highest initial variability, with mean readability at 2.61 and a wide 95\% confidence interval range of 5.86 when rated by two linguists; this range decreased notably to 3.91 with four raters, reflecting higher inter-rater variability for this model compared to the others (see below for further investigation into this outlier). In contrast, ratings for Claude 3.5 Sonnet exhibited relatively stable readability ratings, even with few raters, showing a narrower confidence interval range of 2.60 for two raters, reducing slightly to 2.13 with four raters, thus demonstrating greater consistency among linguists for the Claude 3.5 Sonnet model.

\begin{figure}[tbp]
  \centering
  \includegraphics[width=0.9\linewidth]{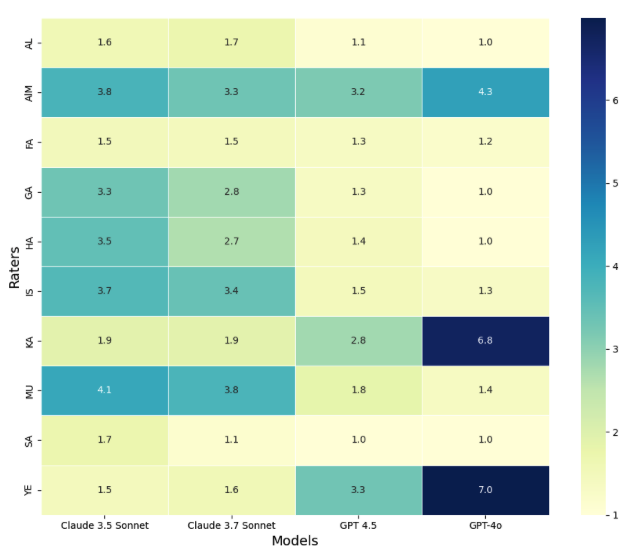}
  \caption{Heatmap of mean Kanuri readability scores by individual linguists. Each cell displays the average score assigned by each linguist rater for sentences generated by different models.}
  \label{fig:synvoices_kanuri_readability_heatmap}
\end{figure}

A heatmap of the mean readability rating agreement among Kanuri raters points to an expected linguist bias and general agreement in model ranking, with the notable exception of GPT-4o (Figure~\ref{fig:synvoices_kanuri_readability_heatmap}). Specifically, three linguists rated GPT-4o highest, with two raters providing mean ratings of 6.6 or higher. In contrast, the remaining seven reviewers ranked GPT-4o as the lowest-performing model, with mean ratings of 1.4 or lower. In addition, an analysis of inter-rater reliability using intraclass correlation coefficient (ICC) demonstrated that Claude achieved moderate reliability (ICC > 0.5) with 5-6 linguists rating 35-50 sentences. In contrast, GPT models showed poor reliability even with 10 linguists, further emphasizing the differences in rater agreement among models (\ref{sec:appendix_irr}). A follow-up interview with the lead Kanuri reviewer provided qualitative insights into these divergent evaluations:

\begin{itemize}
    \item \textbf{GPT-4o}: The sample text was identified as high-quality Hausa, not Kanuri.
    \item \textbf{GPT-4.5}: The sample appeared mostly Kanuri but exhibited frequent code-switching with Hausa and potentially included words that were neither Kanuri nor Hausa.
    \item \textbf{Claude 3.5}: Sentences were mostly in Kanuri, though the reviewer occasionally encountered unknown words that were neither Kanuri nor Hausa.
\end{itemize}

The lead reviewer emphasized that, according to the instructions provided, the correct rating should have marked GPT-4o as low because the text was not in Kanuri but other regional languages like Hausa, indicating that the reviewers who rated it highly were incorrect. We believe this is an interesting finding and recommend that future work investigate methods for improved rater alignment, including further reviewer training or implementing early quality checks on rater consistency, with adjustments as necessary.

\subsubsection{Duplication Challenges for Large Quantity Synthetic Text Generation}

Following the evaluation of synthetic text generation models, we generated large-scale synthetic text corpora for three languages for subsequent use in synthetic voice generation and ASR fine-tuning. The text generation process was identical to that used for text evaluation, except that we utilized a batch processing API for reduced cost. During this large-scale generation, we observed an unexpected large amount of sentence duplication. Specifically, using Claude 3.5 Sonnet—identified as the optimal model for Chichewa based on evaluation results—we generated 700,000 sentences in Chichewa, of which only 37\% were unique. In comparison, text generated using Claude 3.7 Sonnet exhibited significantly less duplication, with 86\% of 530,400 sentences being unique.

\begin{figure}[tbp]
  \centering
  \includegraphics[width=0.9\linewidth]{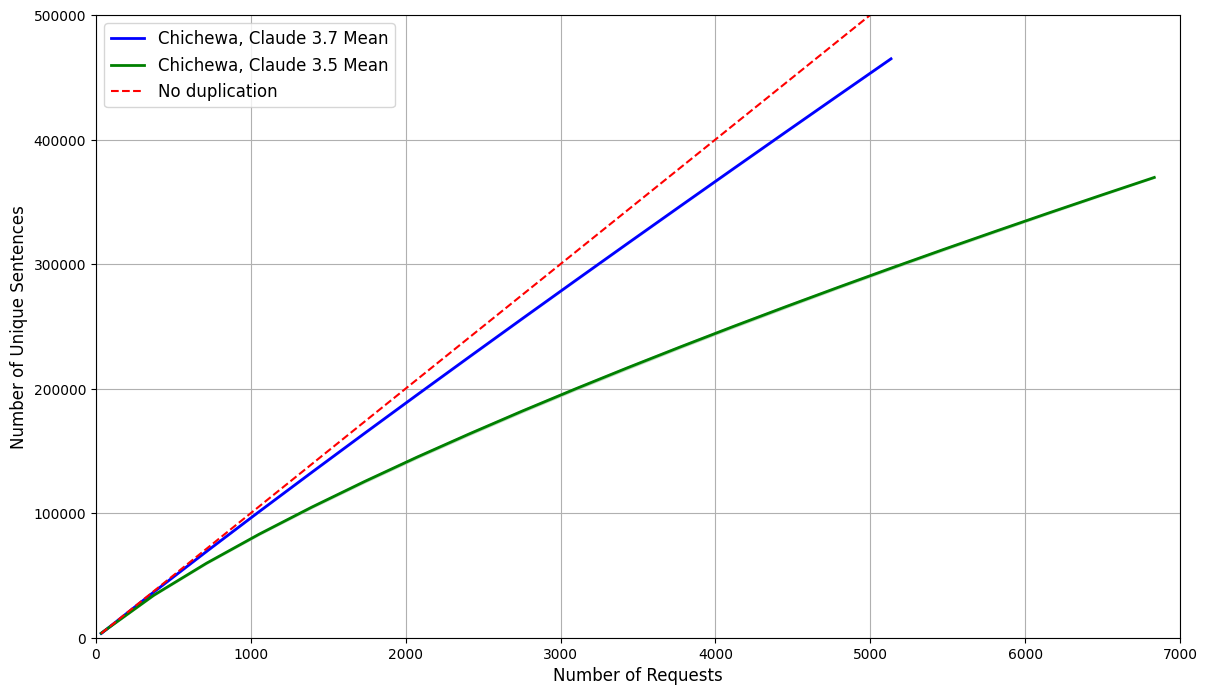}
  \caption{Anthropic Claude unique sentence generation for large-scale Chichewa synthetic text corpora generation.}
  \label{fig:synvoices_claude_duplication}
\end{figure}

To further investigate, we performed a simulation study, subsampling the batch requests without replacement (n=1000 subsamples per observation) to assess the rate of unique sentence generation as a function of batch size (Figure~\ref{fig:synvoices_claude_duplication}). Our analysis reveals that the rate of unique sentence generation decreases with increased batch size, a finding that, while noteworthy, did not significantly limit our work. The deduplicated Chichewa corpus generated by Claude 3.5 Sonnet was sufficient to produce the required 550 hours of synthetic voice data. Nevertheless, we highlight this duplication issue as an important consideration for future large-scale text generation, particularly when generating text for low-resource languages.

\subsection{Synthetic Voice Generation}

Our findings indicate that the TTS model has a substantial impact on synthetic voice data quality, with the best model architectures outperforming the worst by up to 2.03 points for intelligibility and 1.72 points for naturalness on a five-point scale. The VITS-based YourTTS models generally performed better than XTTS, although VITS-based MMS performed better in naturalness for Chichewa (see Table~\ref{tab:tts_eval_results}). We prioritized intelligibility and therefore used the YourTTS model for our synthetic data creation. Our results also indicate that there is not a clearly superior architecture with XTTS outperforming most VITS-based model in Hausa except YourTTS but performed worse than both MMS and YourTTS for Chichewa. 

We find that the quality of the TTS model probably does not matter beyond a certain threshold for the purpose of generating synthetic data. The Hausa ASR models trained on synthetic data generated using YourTTS don’t outperform those trained on data generated by XTTS, even though the former model outperforms the latter on both intelligibility and naturalness (see results below).

\begin{table*}[tb]
\centering
\begin{tabular}{llcccccc}
\toprule
\textbf{Model} & \textbf{Architecture} &
\multicolumn{2}{c}{\textbf{Hausa}} &
\multicolumn{2}{c}{\textbf{Dholuo}} &
\multicolumn{2}{c}{\textbf{Chichewa}} \\
 &  & Intell. & Natural. & Intell. & Natural. & Intell. & Natural. \\
\midrule
MMS & VITS & 2.47 & 2.35 & -- & -- & 4.35 & \textbf{4.03} \\
Original BibleTTS & VITS & 3.53 & 3.53 & -- & -- & -- & -- \\
New BibleTTS & VITS & 3.00 & 2.89 & -- & -- & -- & -- \\
XTTS & Transformer-based & 3.72 & 3.55 & 3.61 & 3.34 & 2.79 & 2.85 \\
YourTTS & VITS & \textbf{4.50} & \textbf{4.07} & \textbf{4.71} & \textbf{4.59} & \textbf{4.45} & 3.82 \\
\bottomrule
\end{tabular}
\caption{Human evaluation of intelligibility and naturalness of different TTS models for Hausa, Dholuo and Chichewa.}
\label{tab:tts_eval_results}
\end{table*}

\subsection{ASR Model Performance with Synthetic Data}

\subsubsection{Medium Data Scenario: Hausa}

\begin{table*}[tb]
\small
\centering
\setlength{\tabcolsep}{6pt}
\renewcommand{\arraystretch}{1.2}
\begin{tabularx}{\textwidth}{l Y Y Y Y Y Y Y Y}
\toprule
\multirow{2}{*}{\textbf{Real-to-synthetic data ratios}} &
\multicolumn{2}{c}{\textbf{FLEURS}} &
\multicolumn{2}{c}{\textbf{NaijaVoices}} &
\multicolumn{2}{c}{\textbf{Common Voice}} \\
 & WER & CER & WER & CER & WER & CER \\
\midrule
\multicolumn{7}{l}{\textbf{500h constant training corpus size}} \\
100h:400h XTTS   & 
28.58 \newline (28.63 $\pm$ 0.86) & 
10.64 \newline (10.04 $\pm$ 0.47) & 
24.57 \newline (24.57 $\pm$ 0.33) & 
6.26 \newline (6.26 $\pm$ 0.12) & 
17.95 \newline (17.94 $\pm$ 0.67) & 
3.73 \newline (3.73 $\pm$ 0.16) \\
100h:400h YourTTS   & 
29.85 \newline (29.88 $\pm$ 0.9) & 
11.57 \newline (11.64 $\pm$ 0.7) & 
27.33 \newline (27.33 $\pm$ 0.34) & 
7.11 \newline (7.11 $\pm$ 0.12) & 
19.8 \newline (19.8 $\pm$ 0.68) & 
4.16 \newline (4.16 $\pm$ 0.17) \\
250h:250h XTTS   & 
26.17 \newline (26.23 $\pm$ 0.65) & 
9.01 \newline (9.03 $\pm$ 0.42) & 
22.91 \newline (22.92 $\pm$ 0.31) & 
5.84 \newline (5.84 $\pm$ 0.17) & 
18.69 \newline (18.68 $\pm$ 0.67) & 
3.67 \newline (3.73 $\pm$ 0.16) \\
250h:250h YourTTS   & 
27.02 \newline (27.03 $\pm$ 0.87) & 
10.47 \newline (10.5 $\pm$ 0.7) & 
23 \newline (22.99 $\pm$ 0.3) & 
5.75 \newline (5.75 $\pm$ 0.11) & 
18.73 \newline (18.74 $\pm$ 0.69) & 
3.64 \newline (3.64 $\pm$ 0.16) \\
\midrule
\multicolumn{7}{l}{\textbf{Real data ablation}} \\
100h:0      & 
30.23 \newline (30.25 $\pm$ 0.81) & 
10.58 \newline (10.6 $\pm$ 0.54) & 
24.06 \newline (24.07 $\pm$ 0.34) & 
6.23 \newline (6.24 $\pm$ 0.12) & 
19.53 \newline (19.52 $\pm$ 0.67) & 
4.03 \newline (4.03 $\pm$ 0.16) \\
250h:0      & 
27.8 \newline (27.79 $\pm$ 0.78)  & 
10.21 \newline (10.19 $\pm$ 0.57) & 
23.01 \newline (23.0 $\pm$ 0.31)  & 
5.75 \newline (5.75 $\pm$ 0.17) & 
19.04 \newline (19.04 $\pm$ 0.67) & 
3.76 \newline (3.76 $\pm$ 0.15) \\
500h:0h &
26.91 \newline (26.9 $\pm$ 0.67) &
9.64 \newline (9.63 $\pm$ 0.47) &
22.49 \newline (22.5 $\pm$ 0.34) &
5.71 \newline (5.72 $\pm$ 0.11) &
17.91 \newline (17.9 $\pm$ 0.67) &
3.60 \newline (3.61 $\pm$ 0.15) \\
\midrule
\multicolumn{7}{l}{\textbf{Full data}} \\
579h:450h XTTS   & 
\textbf{25.73} \newline (25.75 $\pm$ 0.63) & 
\textbf{8.96} \newline (8.97 $\pm$ 0.44) & 
22.43 \newline (22.42 $\pm$ 0.33) & 
5.74 \newline (5.74 $\pm$ 0.11) & 
18.16 \newline (18.17 $\pm$ 0.67) & \
\textbf{3.44} \newline (3.44 $\pm$ 0.15) \\
579h:993h YourTTS   & 
28.42 \newline (28.47 $\pm$ 0.98) & 
11.22 \newline (11.27 $\pm$ 0.79) & 
\textbf{22.06} \newline (22.06 $\pm$ 0.3) & 
\textbf{5.64} \newline (5.64 $\pm$ 0.12) & 
\textbf{17.45} \newline (17.42 $\pm$ 0.66) & 
3.45 \newline (3.45 $\pm$ 0.15) \\
\bottomrule
\end{tabularx}
\caption{WER and CER for Wav2Vec-Bert 2.0 Hausa models trained on different ratios of real and synthetic data. In parentheses, we present bootstrapped mean and standard deviation WER and CER.}
\label{tab:hausa_asr_results}
\end{table*}

Given that the total training corpus size is equal among the 500h:0h, 100h:400h (1:4 real:synthetic) and 250h:250h (1:1 real:synthetic) models, we sought a constant or improving performance to show that real data can be substituted with synthetic data without performance loss, thereby implying that similarly performing models can also be trained on less human data.

While the performance differs between the evaluation datasets, in general, replacing half of the human data with synthetic data, i.e. a 250h:250h ratio, in model training yields performance only marginally worse or even better than a model trained on 500h of real data (see Table~\ref{tab:hausa_asr_results}). Especially models trained with synthetic data created with XTTS perform marginally better. On the CommonVoice testdata, even a model trained on a 1:4 ratio performs better than the model trained on 500h of real data.

We also tested whether these results might stem from model saturation at 100h or 250h of real data. However, the data ablation study indicated that this is not the case as the models still showed the same slight improvements as with adding real data.
Usually, increasing the training corpus yields higher performance, and we observed the same, with our full data model trained on all available data showing the best performance across most evaluation sets and metrics, albeit with only minor improvements.

We also evaluated performance by gender. Given that our synthetic speech data is exclusively male, this is of special importance to ensure that we do not develop ASR models that show gender bias and perform worse for female voices. 

We found that on average the fine-tuned models perform slightly worse for male voices than for female voices (see the ~\ref{sec:gender_disaggregated_asr} for detailed results). On FLEURS, the models performed on average $\sim9.67$ percentage points WER and $\sim0.82$ percentage points CER worse for men than for women. Closer inspection of the Hausa FLEURS evaluation datasets showed that it only contained one male sample, probably explaining these stark differences. NaijaVoices and Common Voice therefore provide better gender-disaggregated evaluation. The gender-disaggregated performance on the NaijaVoices test and Common Voice test set showed an average difference in WER/ CER is $\sim1.82 / \sim{-0.17}$ and $\sim1.29 / \sim0.57$ percentage points, respectively (positive numbers indicate worse performance for male speakers).

\subsubsection{Low Data Scenario: Dholuo and Chichewa}

Dholuo and Chichewa are representative of the majority of African languages that have the same or less speech data available. As 19 and 34 hours, respectively, are not sufficient to train well-functioning ASR models, we added increasing amounts of synthetic data to increase the total training corpus. Therefore, we would expect improvements in performance as the training corpus size increases.

\begin{table*}[tb]
\small
\centering
\setlength{\tabcolsep}{6pt}
\renewcommand{\arraystretch}{1.2}
\begin{tabularx}{\textwidth}{l Y Y Y Y}
\toprule
\multirow{2}{*}{\textbf{Real-to-synthetic data ratios}} &
\multicolumn{2}{c}{\textbf{FLEURS}} &
\multicolumn{2}{c}{\textbf{Common Voice}} \\
 & WER & CER & WER & CER \\
\midrule
19h:0h & 
\textbf{26.92} \newline (26.92 $\pm$ 0.91) & 
6.07 \newline (6.06 $\pm$ 0.29) & 
30.65 \newline (30.64 $\pm$ 0.51) & 
6.99 \newline (6.99 $\pm$ 0.22) \\
19h:19h & 
27.15 \newline (27.22 $\pm$ 0.83) & 
6.43 \newline (6.45 $\pm$ 0.30) & 
\textbf{28.75} \newline (28.76 $\pm$ 0.46) & 
\textbf{6.10} \newline (6.09 $\pm$ 0.15) \\
19h:38h & 
29.4 \newline (29.4 $\pm$ 0.86) & 
6.61 \newline (6.59 $\pm$ 0.30) & 
29.25 \newline (29.27 $\pm$ 0.49) & 
6.55 \newline (6.56 $\pm$ 0.21) \\
19h:77h & 
28.28 \newline (28.26 $\pm$ 0.73) & 
\textbf{6.01} \newline (6.00 $\pm$ 0.23) & 
30.18 \newline (30.2 $\pm$ 0.50) & 
6.69 \newline (6.69 $\pm$ 0.17) \\
\bottomrule
\end{tabularx}
\caption{WER and CER for Wav2Vec-Bert 2.0 Dholuo models trained on different ratios of real and synthetic data. In parentheses, we present bootstrapped mean and standard deviation WER and CER.}
\label{tab:luo_asr_results}
\end{table*}

For Dhuluo, the results depended on the test set (see Table~\ref{tab:luo_asr_results}). On FLEURS, no amount of synthetic data did  improve the WER and improvements in CER were not statistically significant (e.g. the improvement 6.06 to 6.00 CER when adding 77h of synthetic data is well within the standard deviation of 0.29 and 0.23 respectively). On CommonVoice, adding 19h of synthetic data for a 1:1 ratio improved performance from 30.64 $\pm$ 0.51 to 28.76 $\pm$ 0.46 WER and from 6.99 $\pm$ 0.22 to 6.09 $\pm$ 0.15 CER. Adding further synthetic data did not yield further improvements.

\begin{table*}[tb]
\small
\centering
\setlength{\tabcolsep}{6pt}
\renewcommand{\arraystretch}{1.2}
\begin{tabularx}{\textwidth}{l Y Y Y Y}
\toprule
\multirow{2}{*}{\textbf{Real-to-synthetic data ratios}} &
\multicolumn{2}{c}{\textbf{FLEURS}} &
\multicolumn{2}{c}{\textbf{Zambezi Voice}} \\
 & WER & CER & WER & CER \\
\midrule
34h:0h & 
35.38 \newline (35.39 $\pm$ 0.59) & 
7.67 \newline (7.67 $\pm$ 0.40) & 
19.76 \newline (19.76 $\pm$ 0.70) & 
4.51 \newline (4.52 $\pm$ 0.40) \\
34h:34h & 
34.32 \newline (34.33 $\pm$ 0.59) & 
7.56 \newline (7.55 $\pm$ 0.39) & 
19.90 \newline (19.86 $\pm$ 0.74) & 
4.57 \newline (4.56 $\pm$ 0.39) \\
34h:68h & 
33.39 \newline (33.4 $\pm$ 0.59) & 
\textbf{7.15} \newline (7.15 $\pm$ 0.36) & 
\textbf{18.53} \newline (18.54 $\pm$ 0.71) & 
\textbf{4.38} \newline (4.41 $\pm$ 0.38) \\
34h:102h & 
34.10 \newline (34.1 $\pm$ 0.61) & 
7.42 \newline (7.43 $\pm$ 0.41) & 
20.28 \newline (20.3 $\pm$ 0.76) & 
4.74 \newline (4.75 $\pm$ 0.40) \\
34h:136h & 
34.72 \newline (34.71 $\pm$ 0.59) & 
7.65 \newline (7.65 $\pm$ 0.40) & 
21.20 \newline (21.21 $\pm$ 0.72) & 
4.96 \newline (4.97 $\pm$ 0.39) \\
34h:307h & 
\textbf{32.95} \newline (32.95 $\pm$ 0.61) & 
7.27 \newline (7.25 $\pm$ 0.38) & 
18.69 \newline (18.71 $\pm$ 0.70) & 
4.46 \newline (4.48 $\pm$ 0.38) \\
\bottomrule
\end{tabularx}
\caption{WER and CER for Wav2Vec-Bert 2.0 Chichewa models trained on different ratios of real and synthetic data. In parentheses, we present bootstrapped mean and standard deviation WER and CER.}
\label{tab:chichewa_asr_results}
\end{table*}

In contrast, for Chichewa (see Table~\ref{tab:chichewa_asr_results}), we found consistent improvements when adding synthetic data. Adding 68 hours of synthetic data for a 1:2 ratio between real and synthetic data and adding 307h of synthetic data for a 1:9 ratio resulted in the best performing models. On Zambezi Voice, the 1:2 ratio yielded the best absolute performance of 18.54 $\pm$ 0.71 WER and 4.41 $\pm$ 0.38 CER although this is not statistically better than the performance of a 1:9 ratio. Evaluation on the FLEURS test set confirmed these results, only that the 1:9 ratio exhibited best absolute WER performance

Unfortunately, the available evaluation data did not allow an analysis of performance by gender. For Dholuo, all speakers in the FLEURS and Common Voice evaluation sets are women. For Chichewa, all speakers in the FLEURS evaluation set are male, while Zambezi Voice does not provide gender metadata.

\subsubsection{Evaluation Challenges Due to Non-standardized Scripts and Potential Errors in Evaluation Data}

Spot checks of the errors by different models indicated that part of the word error rate might be due to non-standardized scripts and diacritics in Hausa, Dholuo, and Chichewa, where different but equally legitimate ways of writing the same word are counted as errors and where diacritics are not consistently used or transcribed. This aligns with work on potential limitations of WER \citep{aksenova-etal-2021-might} and benchmarking for Indic languages \citep{watts2024parikshalargescaleinvestigationhumanllm}. 
 
Although a thorough analysis of this issue is beyond the scope of this paper, we used the NeMo Speech Data Explorer \citep{bakhturina2021a} to extract some of the unrecognized words with the highest frequency in our evaluation set transcriptions. These are words present in the evaluation transcript, but which the ASR model always fails to transcribe correctly.

Our first results created by native speakers evaluating selected errors from the various models indicate that we potentially underestimate the ASR performance. Of 19 to 20 errors per language evaluated, the evaluators labeled all 19 as no errors for Dholuo, 20 as no errors for Hausa (with 4 errors in the evaluation transcript), and 2 out of 20 as no errors for Chichewa. We present examples to illustrate challenges and potential shortcomings of current evaluation datasets and metrics. Further examples can be found in ~\ref{sec:human_asr_eval}. Those results also imply that comparisons of WER between languages are not robust, as those issues differ between languages in kind and number. Text normalization as a potential remedy is an ongoing field of research. However, some research has shown that current techniques might not be appropriate for low-resource languages with non-Latin scripts \citep{manohar2024lostnormalizationexploringpitfalls}.

\begin{table*}[p]
\centering
\renewcommand{\arraystretch}{1.3}
\begin{tabularx}{\textwidth}{l X X X X}
\toprule
\textbf{Language} &
\textbf{Evaluation transcript} &
\textbf{Model output} &
\textbf{Evaluator assessment} &
\textbf{Evaluator comments} \\
\midrule
Hausa &
sautin dala da wasan haske na daya daga cikin abubuwa masu dadi a fanni kananan yara &
sautin dala da wasan haske na ɗaya daga cikin abubuwa masu daɗi a fanni ƙananan yara &
No error &
The only difference is the use of special Hausa characters. \\
\midrule
Hausa &
za'a yi mata aiki a ƙwaƙwalwa &
za a yi mata aiki a ƙwaƙwalwa &
Error in evaluation transcript &
Grammatically, the correct form is 'za a' not 'za'a' However, many people are using 'za'a'. \\
\midrule
Hausa &
mallam aminu dan kasuwa ne 'à' kasuwan kure &
malam aminu ɗan kasuwa ne a kasuwan kure &
Error in original transcript (see Abdulminim et al.\ 2024 for known errors in FLORES) &
Model output is the correct version. There is no à in Hausa writing style that we physically see and read. \\
\midrule
Dholuo &
otho sa adek okinyi &
otho saa adek okinyi &
No error &
No error, we either say 'sa' or 'saa'. \\
\midrule
Dholuo &
pesa jaduong ile &
pesa jaduong’ ile &
No error &
Optional final apostrophe. \\
\midrule
Chichewa &
kujambula makanema kwabweletsa zofunikira pakumasulira kwa mawonedwe ang'onoang'ono ofotokoza chinachake kuyenda kwa nkhope komwe kumatha kutenga kanthawi kochepa kwambiri zedi monga ma millisecond ochepa &
kujambula makanema kwabweretsa zofunikira pakumasulira kwa maonedwe ang'onoang'ono ofotokoza chinachake kuyenda kwa nkhope komwe kumatha kutenga kanthawi kochepa kwambiri zedi monga milceand ochepa &
Errors in both original transcript and model output &
\textbf{Evaluation transcription}: \textit{Milisecond} should be transliterated to \textit{Millisekondi}. 
\textbf{Model Output}: \textit{milceand} is a wrong spelling of millisecond, which is normally transliterated as \textit{milisekondi}. \\
\midrule
Chichewa &
ku barcelena chiyankhulo chovomelezeka ndi catalan ndichi sipanishi theka la anthu amakonda kuyakhula catalan ambiri amachimvetsesa ndipo pafupifupi onse amamva ndikudziwa chi sipanishi &
ku barcelona chiyankhulo chovomelezeka ndi catalan ndi chi spanishi thekalathu amakonda kuyankhula catalan ambiri amachimvetsetsa ndipo pafupifupi onse amava ndikudziwa chi spanishi &
Errors in both original transcript and model output &
\textbf{Evaluation transcription}: \textit{ndi} should not be combined with \textit{chi} but rather goes together with the name of the language to read: \textit{chisipanishi}. \textit{Chi} and \textit{sipanishi} should be written conjunctively.

\textbf{Model Output}: \textit{Chi} and \textit{sipanishi} should be written conjunctively. \\
\bottomrule
\end{tabularx}
\caption{Human evaluation of ASR model errors across Hausa, Dholuo, and Chichewa.}
\label{tab:human_asr_eval}
\end{table*}

\subsubsection{Limitations}

Our work is limited to the selected languages, and future research would need to extend the language coverage of studies on synthetic data for African languages. In addition, we could only explore a certain set of parameters for our data generation pipeline and model training. As illustrated in the previous sections, our results are also limited by potential issues in the evaluation datasets that we used, despite their common usage. Furthermore, we present our findings on challenges in human evaluation for low-resource languages, which we think require further investigation.

\section{Conclusion and Future Work}
We investigated the creation of synthetic text and voice data for 10 African languages. Our results show that synthetic text generation with LLMs is feasible for various languages, except for the lowest-resourced languages such as Kanuri or Kinande. We estimate that we created high-quality synthetic voice data for Hausa, Dholuo and Chichewa for less than 1\% the cost of human data\footnote{These costs only include LLM API costs and GPU costs for TTS training and data creation. Costs might be higher if data for TTS model training needs to be generated but likely will remain substantially lower. In addition, as a minimum amount of real data is usually needed, this data might also serve as training data for the ASR model, although we did not explore this process.}, based on internal estimates of the cost of real voice data. Our results show promising utility of the so-created synthetic voice data in complementing human data when training ASR models. But our results also indicate that a minimum of human data is needed. For Hausa, we show that the use of synthetic data either worsens the performance for male voices or does not increase gender bias in ASR performance, depending on the evaluation dataset and in spite of our synthetic data only including male voices. Further investigations also illustrated the challenges of working with human evaluators in low-resource languages where code-mixing and non-standardized scripts are common, as well as the limitations and shortcomings of existing evaluation datasets and resulting metrics.

\subsection{Avenues for future research}

Our research could show the utility of synthetic voice data in a controlled setting on commonly used evaluation sets. Future research should further investigate the robustness of synthetic data for use in practical applications. This might include investigating the utility of multiple-speaker TTS and voice cloning based on very small voice samples to create more diverse synthetic data \citep{ogun20241000africanvoicesadvancing, yang2024enhancinglowresourceasrversatile}. This might allow for the creation of synthetic data for specific use cases and speaker groups. 

\citet{yang2024enhancinglowresourceasrversatile} show that text diversity plays a key role in the utility of synthetic voice data. \citet{chen2024diversitysyntheticdataimpact} and \citet{finch-choi-2024-diverse} show the positive effects of synthetic text diversity for LLM training and dialogue state tracking. Therefore, future work should investigate methods and effectiveness of increasing text diversity and options to target text generation to specific domains and use cases (see \citet{zhu2025measuringdiversitysyntheticdatasets} on how to measure text diversity in synthetic data).

Our work illustrates the challenges in human evaluation. Intercoder reliability can be low and should be monitored. Improved evaluation guidelines are likely needed to increase reliability. We showed that unreliable text generation, e.g. in cases where LLMs create text in better-resourced languages like Hausa when prompted for very low-resource languages like Kanuri, further aggravates intercoder reliability. Further work on synthetic text data would likely benefit from the investigation of which metrics are most suitable and indicative of downstream performance.

The examples presented illustrate that beyond synthetic data, ASR evaluation in low-resource languages requires further investigation to handle non-standardized scripts, either through semantic measures, measures robust to plurality in spellings or language-appropriate normalizers, and work on improved evaluation datasets.

Lastly, further work should be undertaken to investigate other uses of synthetic data in African languages, including the development of (small) language models and N-gram language models to further improve ASR models.

\section{Acknowledgements}
CLEAR Global is grateful for the support of the Gates Foundation that enabled this work and the sub-award and partnership with Dimagi. The authors are indebted to Polly Harlow and Arisha Siddiqui, whose organizational skills we could not replace. We are also grateful to Daniel Wilson at XRI Global, Muhammad Abdul-Mageed and his team at the University of British Columbia, and Howard Lakougna at the Gates Foundation, who provided trusted partnership and valuable feedback. We thank Alp Öktem at CLEAR Global for review and feedback and Joyce Nabende, Alvin Nahabwe, and the team at Makerere University for the close collaboration and exchange around their related project on ASR in African languages, which informed and strengthened our work. Lastly, we would like to thank the evaluators from the TWB Community, without whom we could not have implemented this research.

This work was supported by the Gates Foundation (Grant number INV-076358). The conclusions and opinions expressed in this work are those of the authors alone and shall not be attributed to the Foundation.

\bibliographystyle{plainnat}
\bibliography{references}  

\clearpage
\appendix
\clearpage
\appendix
\renewcommand{\thesection}{Appendix \Alph{section}}

\begin{landscape}
\section{Synthetic text generation language evaluation summary statistics}
~\label{sec:appendix_syntextsummary}

\begin{table}[h]
\centering
\renewcommand{\arraystretch}{1.3}
\begin{tabularx}{\linewidth}{l X X X X X X}
\toprule
\textbf{Language} &
\textbf{Model} &
\textbf{Readability \&\newline Naturalness [1..7]} &
\textbf{Grammatical \newline Correctness [0,1]} &
\textbf{Real Words [0,1]} &
\textbf{Notable Error [0,1]} &
\textbf{Adequacy \& \newline Accuracy [1..7]} \\
\midrule
\multirow{4}{*}{Bambara} &
Claude 3.5 Sonnet &
6.08 $\pm$ 1.18 &
0.82 $\pm$ 0.38 &
0.82 $\pm$ 0.38 &
0.21 $\pm$ 0.41 &
5.83 $\pm$ 1.41 \\
&
Claude 3.7 Sonnet &
5.80 $\pm$ 1.26 &
0.78 $\pm$ 0.41 &
0.78 $\pm$ 0.42 &
0.25 $\pm$ 0.43 &
5.63 $\pm$ 1.43 \\
&
GPT-4o &
3.72 $\pm$ 2.08 &
0.20 $\pm$ 0.40 &
0.31 $\pm$ 0.46 &
0.87 $\pm$ 0.34 &
2.54 $\pm$ 1.34 \\
&
GPT-4.5 &
3.95 $\pm$ 1.84 &
0.24 $\pm$ 0.43 &
0.43 $\pm$ 0.50 &
0.78 $\pm$ 0.41 &
2.85 $\pm$ 1.40 \\
\midrule
\multirow{4}{*}{Chichewa} &
Claude 3.5 Sonnet &
5.88 $\pm$ 1.05 &
0.75 $\pm$ 0.43 &
0.87 $\pm$ 0.33 &
0.13 $\pm$ 0.33 &
4.62 $\pm$ 1.81 \\
&
Claude 3.5 Sonnet* &
5.20 $\pm$ 1.77 &
0.67 $\pm$ 0.47 &
0.92 $\pm$ 0.28 &
0.27 $\pm$ 0.44 &
4.91 $\pm$ 1.62 \\
&
GPT-4o &
5.73 $\pm$ 1.11 &
0.76 $\pm$ 0.43 &
0.91 $\pm$ 0.29 &
0.19 $\pm$ 0.40 &
4.54 $\pm$ 1.66 \\
&
O1 &
5.00 $\pm$ 1.72 &
0.59 $\pm$ 0.49 &
0.84 $\pm$ 0.37 &
0.25 $\pm$ 0.44 &
4.02 $\pm$ 1.62 \\
\midrule
\multirow{2}{*}{Hausa} &
Claude 3.5 Sonnet &
5.14 $\pm$ 1.26 &
0.38 $\pm$ 0.49 &
0.47 $\pm$ 0.50 &
0.63 $\pm$ 0.48 &
5.07 $\pm$ 1.39 \\
&
GPT-4o &
5.67 $\pm$ 0.94 &
0.59 $\pm$ 0.49 &
0.64 $\pm$ 0.48 &
0.42 $\pm$ 0.49 &
5.63 $\pm$ 1.03 \\
\midrule
\multirow{5}{*}{Kanuri} &
Claude 3.5 Sonnet &
2.67 $\pm$ 1.64 &
0.02 $\pm$ 0.14 &
0.14 $\pm$ 0.34 &
0.58 $\pm$ 0.49 &
1.38 $\pm$ 0.98 \\
&
Claude 3.7 Sonnet &
1.18 $\pm$ 0.57 &
0.00 $\pm$ 0.00 &
0.00 $\pm$ 0.07 &
1.00 $\pm$ 0.00 &
1.33 $\pm$ 0.64 \\
&
GPT-4o &
1.55 $\pm$ 0.70 &
0.02 $\pm$ 0.14 &
0.00 $\pm$ 0.00 &
0.51 $\pm$ 0.50 &
1.40 $\pm$ 1.39 \\
&
GPT-4.5 &
2.02 $\pm$ 2.10 &
0.11 $\pm$ 0.32 &
0.11 $\pm$ 0.32 &
0.91 $\pm$ 0.29 &
2.03 $\pm$ 2.00 \\
&
O1 &
1.51 $\pm$ 0.64 &
0.00 $\pm$ 0.00 &
0.00 $\pm$ 0.00 &
0.51 $\pm$ 0.50 &
1.03 $\pm$ 0.30 \\
\midrule
\multirow{3}{*}{Dholuo} &
Claude 3.5 Sonnet &
5.91 $\pm$ 1.42 &
0.78 $\pm$ 0.42 &
0.94 $\pm$ 0.24 &
0.48 $\pm$ 0.50 &
5.82 $\pm$ 1.48 \\
&
GPT-4o &
3.45 $\pm$ 2.01 &
0.15 $\pm$ 0.36 &
0.85 $\pm$ 0.36 &
0.91 $\pm$ 0.29 &
3.22 $\pm$ 2.01 \\
&
O1 &
3.64 $\pm$ 2.09 &
0.22 $\pm$ 0.42 &
0.80 $\pm$ 0.40 &
0.87 $\pm$ 0.34 &
3.23 $\pm$ 1.99 \\
\bottomrule
\end{tabularx}
\end{table}
\end{landscape}

\begin{landscape}
\vspace*{1cm}
\begin{table}
\centering
\renewcommand{\arraystretch}{1.3}
\begin{tabularx}{\linewidth}{l X X X X X X}
\toprule
\textbf{Language} &
\textbf{Model} &
\textbf{Readability \&\newline Naturalness [1..7]} &
\textbf{Grammatical \newline Correctness [0,1]} &
\textbf{Real Words [0,1]} &
\textbf{Notable Error [0,1]} &
\textbf{Adequacy \& \newline Accuracy [1..7]} \\
\midrule
\multirow{4}{*}{Kinande} &
Claude 3.5 Sonnet &
3.61 $\pm$ 1.77 &
0.26 $\pm$ 0.44 &
0.31 $\pm$ 0.46 &
0.79 $\pm$ 0.41 &
2.99 $\pm$ 1.73 \\
&
Claude 3.7 Sonnet &
3.38 $\pm$ 1.61 &
0.22 $\pm$ 0.41 &
0.28 $\pm$ 0.45 &
0.79 $\pm$ 0.41 &
2.86 $\pm$ 1.57 \\
&
GPT-4o &
1.77 $\pm$ 0.96 &
0.01 $\pm$ 0.10 &
0.01 $\pm$ 0.10 &
0.98 $\pm$ 0.14 &
1.58 $\pm$ 0.94 \\
&
GPT-4.5 &
2.51 $\pm$ 1.31 &
0.08 $\pm$ 0.26 &
0.10 $\pm$ 0.30 &
0.90 $\pm$ 0.30 &
2.16 $\pm$ 1.27 \\
\midrule
\multirow{3}{*}{Northern Somali} &
Claude 3.5 Sonnet &
6.06 $\pm$ 1.28 &
0.485 $\pm$ 0.50 &
0.94 $\pm$ 0.24 &
0.33 $\pm$ 0.47 &
6.37 $\pm$ 1.03 \\
&
GPT-4o &
6.03 $\pm$ 1.35 &
0.78 $\pm$ 0.41 &
0.97 $\pm$ 0.17 &
0.19 $\pm$ 0.39 &
6.50 $\pm$ 0.97 \\
&
O1 &
5.97 $\pm$ 1.43 &
0.47 $\pm$ 0.50 &
0.95 $\pm$ 0.23 &
0.30 $\pm$ 0.46 &
6.33 $\pm$ 1.06 \\
\midrule
\multirow{4}{*}{Twi} &
Claude 3.5 Sonnet &
5.53 $\pm$ 1.50 &
0.62 $\pm$ 0.49 &
0.71 $\pm$ 0.45 &
0.54 $\pm$ 0.50 &
5.18 $\pm$ 1.91 \\
&
Claude 3.7 Sonnet &
5.49 $\pm$ 1.51 &
0.40 $\pm$ 0.49 &
0.52 $\pm$ 0.50 &
0.62 $\pm$ 0.48 &
4.74 $\pm$ 1.94 \\
&
GPT-4o &
4.67 $\pm$ 1.81 &
0.49 $\pm$ 0.50 &
0.81 $\pm$ 0.39 &
0.71 $\pm$ 0.45 &
4.10 $\pm$ 2.04 \\
&
O1 &
4.35 $\pm$ 1.91 &
0.38 $\pm$ 0.49 &
0.63 $\pm$ 0.48 &
0.73 $\pm$ 0.44 &
3.87 $\pm$ 2.11 \\
\midrule
\multirow{3}{*}{Wolof} &
Claude 3.5 Sonnet &
5.77 $\pm$ 0.93 &
0.93 $\pm$ 0.25 &
0.66 $\pm$ 0.47 &
0.31 $\pm$ 0.46 &
5.97 $\pm$ 1.35 \\
&
GPT-4o &
5.04 $\pm$ 1.54 &
0.97 $\pm$ 0.17 &
0.82 $\pm$ 0.39 &
0.65 $\pm$ 0.48 &
4.94 $\pm$ 1.77 \\
&
O1 &
5.04 $\pm$ 1.41 &
0.97 $\pm$ 0.18 &
0.72 $\pm$ 0.45 &
0.25 $\pm$ 0.43 &
4.96 $\pm$ 1.66 \\
\midrule
\multirow{3}{*}{Yoruba} &
Claude 3.5 Sonnet &
6.14 $\pm$ 0.86 &
0.93 $\pm$ 0.26 &
0.96 $\pm$ 0.19 &
0.25 $\pm$ 0.43 &
5.92 $\pm$ 1.16 \\
&
GPT-4o &
5.59 $\pm$ 1.37 &
0.68 $\pm$ 0.47 &
0.97 $\pm$ 0.18 &
0.46 $\pm$ 0.50 &
5.54 $\pm$ 1.43 \\
&
O1 &
5.35 $\pm$ 1.38 &
0.71 $\pm$ 0.46 &
0.90 $\pm$ 0.30 &
0.54 $\pm$ 0.50 &
5.26 $\pm$ 1.51 \\
\bottomrule
\end{tabularx}
\end{table}
\end{landscape}

\clearpage 

\renewcommand{\thesection}{Appendix \Alph{section}}
\renewcommand{\thefigure}{B.\arabic{figure}}
\setcounter{figure}{0} 

\twocolumn[
\section{Synthetic Kanuri text inter-rater reliability analysis}
~\label{sec:appendix_irr}
\vspace{1em}
]

This appendix analyzes inter-rater reliability for 400 sentences generated in Kanuri using the methodology described in Section ~\ref{sec:methodology_text}. The corpus is sourced equally from four LLMs: Claude 3.5 Sonnet, Claude 3.7 Sonnet, GPT-4o and GPT-4.5. Ten native-speaking linguists performed a blind evaluation of the 400 randomly shuffled sentences on three metrics meant to capture language quality in the target language: readability and naturalness of the sentence, grammatical correctness and all words being from the target language. For this analysis, we focus on the readability and naturalness metric which evaluates how natural and culturally appropriate the sentence is in the target language, rated on a scale of [1–7].

Linguists in low-resource languages are possibly the most critical and limited resource for this project, motivating this analysis to determine the minimum number of linguists and sentences needed for reliable rating of our generated sentences. We measure linguists' agreement of sentence readability in Kanuri using the intraclass correlation coefficient (see \citep{shrout1979}). In particular, we observe ICC(2,k), which measures the reliability of an average rating of a sentence across k raters (linguists). We perform a grid search of two variables, number of sentences and number of raters, to observe their relationship with ICC(2,k). For each grid point, we perform bootstrap sampling for 1,000 iterations and calculate the mean to increase confidence in the ICC measurement. According to \citep{koo2016}, ICC values can be interpreted as follows: "values less than 0.5, between 0.5 and 0.75, between 0.75 and 0.9, and greater than 0.90 are indicative of poor, moderate, good, and excellent reliability, respectively."

\begin{figure}[tbp]
    \centering
    \begin{subfigure}{0.45\textwidth}
        \centering
        \includegraphics[width=0.9\linewidth]{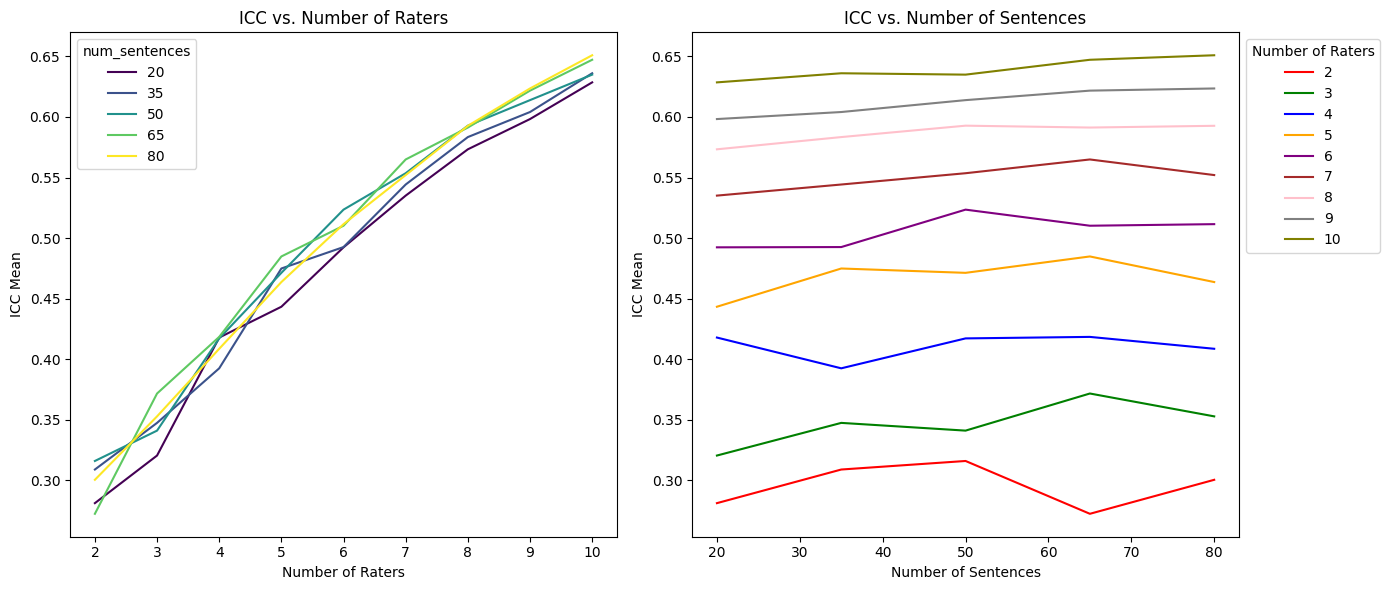}
        \caption{}
    \end{subfigure}
    \hfill
    \begin{subfigure}{0.45\textwidth}
        \centering
        \includegraphics[width=0.9\linewidth]{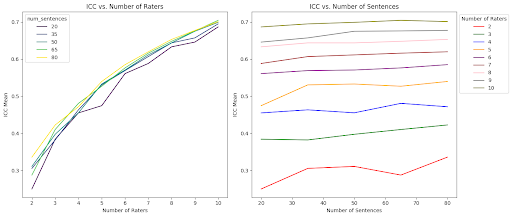}
        \caption{}
    \end{subfigure}
        
    \begin{subfigure}{0.45\textwidth}
        \centering
        \includegraphics[width=0.9\linewidth]{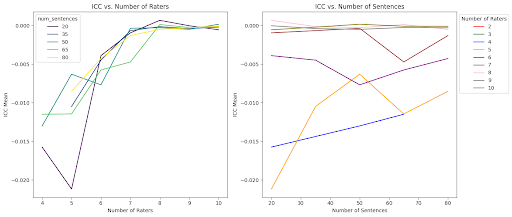}
        \caption{}
    \end{subfigure}
    \hfill
    \begin{subfigure}{0.45\textwidth}
        \centering
        \includegraphics[width=0.9\linewidth]{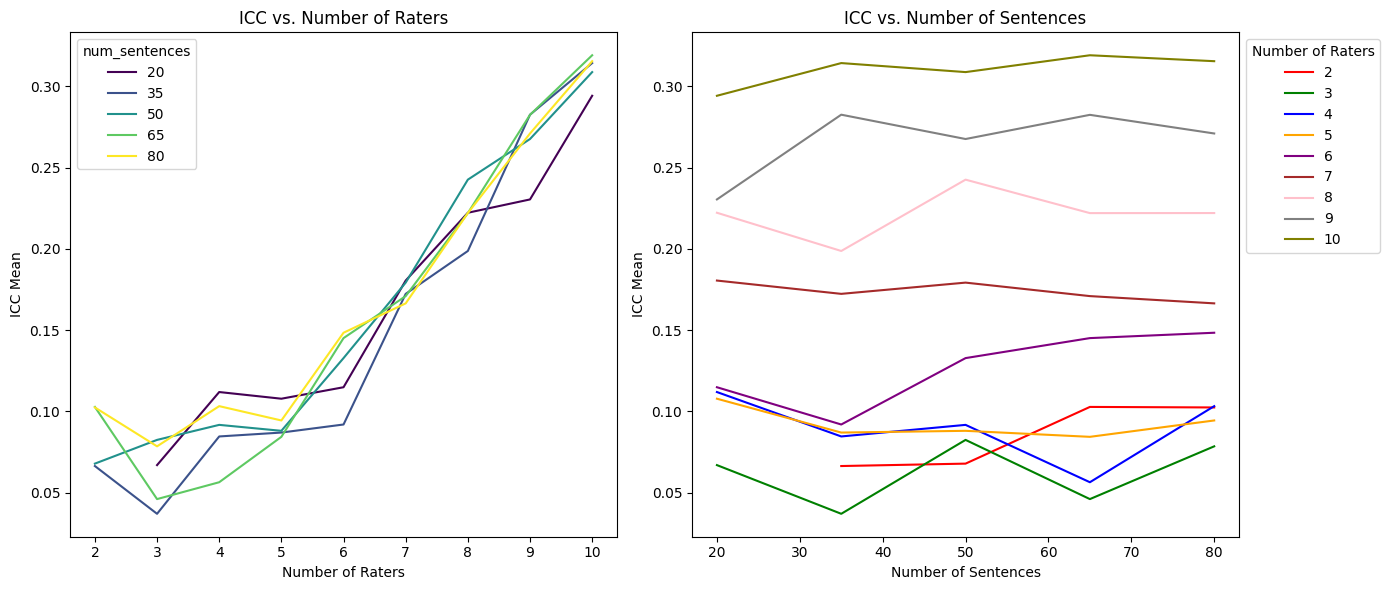}
        \caption{}
    \end{subfigure}
    
    \caption{Observed mean ICC(2,k) for a varying number of raters and sentences rated in Kanuri for (a) Claude 3.5 Sonnet, (b) Claude 3.7 Sonnet, (c) GPT-4o and (d) GPT4.5.}
    \label{fig:kanuri_irr}
\end{figure}

Figure ~\ref{fig:kanuri_irr} shows the impact of sentence volume and number of linguists on ICC(2,k) for each LLM. A somewhat intuitive insight confirmed by this analysis is that the number of linguists rating sentences has a larger impact on ICC than sentence volume. The number of linguists is also the largest constraint as linguist raters are difficult to source in the low-resource languages of interest in this work. For Claude 3.5 and Claude 3.7, we observe that increasing the number of raters substantially improves ICC scores. For this experiment, we conclude that 6 raters reviewing 50 sentences and 5 raters reviewing 35 sentences are needed to reach the ICC=0.5 moderate threshold for Claude 3.5 and Claude 3.7, respectively. For the OpenAI models (GPT-4o and GPT-4.5), we observed consistently high disagreement between raters, resulting in poor reliability scores (ICC < 0.5) even with the maximum number of raters and sentences tested. Section ~\ref{sec:results_irr} details the investigation and implications of the linguist rater disagreement.

\renewcommand{\thesection}{Appendix \Alph{section}}
\renewcommand{\thefigure}{C.\arabic{figure}}
\setcounter{figure}{0} 
\renewcommand{\thetable}{C.\arabic{table}}
\setcounter{table}{0} 

\clearpage 
\onecolumn
\section{MMS-1B Performance for Hausa across different ratios of real and synthetic data}
~\label{sec:appendix_mms_asr_results}

\begin{table*}[ht]
\centering
\begin{tabular}{lcccccc}
\toprule
\textbf{Real-to-synthetic data ratios} & \multicolumn{2}{c}{\textbf{FLEURS}} & \multicolumn{2}{c}{\textbf{NaijaVoices}} & \multicolumn{2}{c}{\textbf{Common Voice}} \\
& WER & CER & WER & CER & WER & CER \\
\midrule
50h:0 & 30.34 & 10.45 & \textbf{34.72} & \textbf{9.00} & 27.26 & 5.67 \\
500h:0 & \textbf{29.13} & \textbf{9.19} & 34.92 & 9.02 & \textbf{27.22} & \textbf{5.62} \\
250h:250h XTTS & 29.77 & 10.16 & 37.61 & 9.63 & 28.43 & 5.87 \\
100h:400h XTTS & 30.27 & 9.86 & 38.95 & 10.12 & 27.42 & 5.91 \\
50h:450h XTTS & 31.88 & 10.82 & 40.42 & 10.52 & 28.92 & 6.15 \\
\bottomrule
\end{tabular}
\label{tab:mms_asr_results}
\end{table*}

\renewcommand{\thesection}{Appendix \Alph{section}}
\renewcommand{\thefigure}{D.\arabic{figure}}
\setcounter{figure}{0} 
\renewcommand{\thetable}{D.\arabic{table}}
\setcounter{table}{0} 

\section{Hausa Wav2Vec-BERT 2.0 ASR Results by gender}
~\label{sec:gender_disaggregated_asr}

\begin{table}[ht]
\centering
\renewcommand{\arraystretch}{1.2}
\begin{tabular}{lcc|cc|cc|cc|cc|cc}
\hline
\textbf{Real:Synth Ratio} &
\multicolumn{4}{c|}{\textbf{FLEURS}} &
\multicolumn{4}{c|}{\textbf{NaijaVoices}} &
\multicolumn{4}{c}{\textbf{Common Voice}} \\
\cline{2-13}
& \multicolumn{2}{c|}{Male} & \multicolumn{2}{c|}{Female} 
& \multicolumn{2}{c|}{Male} & \multicolumn{2}{c|}{Female}
& \multicolumn{2}{c|}{Male} & \multicolumn{2}{c}{Female} \\
& \multicolumn{2}{c|}{(n=1)} & \multicolumn{2}{c|}{(n=620)} 
& \multicolumn{2}{c|}{(n=2845)} & \multicolumn{2}{c|}{(n=1679)}
& \multicolumn{2}{c|}{(n=180)} & \multicolumn{2}{c}{(n=34)} \\
& WER & CER & WER & CER 
& WER & CER & WER & CER 
& WER & CER & WER & CER \\
\hline
500h:0     & 28.57 & 8.57  & 26.91 & 9.64  & 23.11 & 5.63 & 21.43 & 5.86 & 19.23 & 3.85 & 13.19 & 2.45 \\
100h:400h XTTS  & 42.86 & 12.86 & 28.57 & 10.64 & 25.36 & 6.28 & 23.24 & 6.23 & 17.35 & 3.82 & 14.47 & 2.80 \\
100h:400h YourTTS  & 42.86 & 15.71 & 29.84 & 11.57 & 28.47 & 7.15 & 25.41 & 7.03 & 19.58 & 4.20 & 21.70 & 4.29 \\
250h:250h XTTS  & 35.71 & 14.29 & 26.16 & 9.00  & 23.35 & 5.71 & 22.16 & 6.05 & 18.47 & 3.68 & 17.87 & 3.68 \\
250h:250h YourTTS  & 35.71 & 10.00 & 27.01 & 10.47  & 23.42 & 5.64 & 22.28 & 5.93 & 19.72 & 3.89 & 14.04 & 2.63 \\
579h:450h XTTS  & 42.86 & 11.43 & 25.71 & 8.96  & 23.06 & 5.68 & 21.39 & 5.85 & 17.70 & 3.46 & 13.62 & 2.80 \\
579h:993h YourTTS  & 35.71 & 10.00 & 28.41 & 11.22  & 22.54 & 5.51 & 21.26 & 5.85 & 17.77 & 3.60 & 17.45 & 3.06 \\
\hline
\end{tabular}
\caption{Gender-disaggregated Hausa WER and CER scores for FLEURS, NaijaVoices, and Common Voice test sets across different real-to-synthetic training ratios.}
\label{tab:gender_disaggregated_asr}
\end{table}

\renewcommand{\thesection}{Appendix \Alph{section}}
\renewcommand{\thefigure}{E.\arabic{figure}}
\setcounter{figure}{0} 
\renewcommand{\thetable}{E.\arabic{table}}
\setcounter{table}{0} 

\clearpage 
\section{Human evaluation of ASR model errors}
~\label{sec:human_asr_eval}

\begin{longtable}{p{0.25\textwidth} p{0.25\textwidth} p{0.15\textwidth} p{0.3\textwidth}}
\caption{Human evaluation of ASR model errors in Hausa.} \\
\label{tab:hausa_human_asr_eval} \\
\toprule
\textbf{Evaluation transcript} &
\textbf{Model output} &
\textbf{Evaluator assessment} &
\textbf{Evaluator comments} \\
\midrule
\endfirsthead
\caption[]{(continued) Human evaluation of ASR model errors in Hausa.} \\
\toprule
\textbf{Evaluation Transcription} & 
\textbf{Model output} & 
\textbf{Evaluator assessment} & 
\textbf{Evaluator comment} \\
\midrule
\endhead

\midrule
\multicolumn{4}{r}{{Continued on next page}} \\
\midrule
\endfoot

\bottomrule
\endlastfoot

sautin dala da wasan haske na daya daga cikin abubuwa masu dadi a fanni kananan yara &
sautin dala da wasan haske na ɗaya daga cikin abubuwa masu daɗi a fanni ƙananan yara &
No error &
The only difference is the use of special Hausa characters. \\
\midrule
a wasu wurare minti daya ya isa ruwa ya tafasa amma a wasu wuraren kuma yana bukatar mintuna da yawa &
a wasu wurare minti ɗaya ya isa ruwa ya tafasa amma a wasu wuraren kuma yana buƙatar mintuna da yawa &	
No error &
The only difference is the use of special Hausa characters. \\
\midrule
a sauran biranen kasar italiya da kuma sauran kasashen duniya musamman a poland an kafa makamancin ginuwar wanda ya samu dubiyar jama'a da dama &
a sauran biranen ƙasar italiya da kuma sauran ƙasashen duniya musamman a foland an kafa makamancin ginuwar wanda ya samu dubiyar jama'a da dama &
No error &
The only difference is the use of special Hausa characters. \\
\midrule
aukuwar tsananin yanayin yanki da na lokacin sun hada da guguwar iska hadari mai dusar kankara guguwar kankara da guguwar ƙura &
aukuwar tsananin yanayin yanki da na lokacin sun haɗa da guguwar iska hadari mai dusar ƙanƙara guguwar ƙanƙara da guguwar kura &
No error &
The only difference is the use of special Hausa characters. \\
\midrule
garken zaki sun kunshi maza manya daya zuwa uku masu dangantaka tare da mata da dama har zuwa talatin tare da 'ya'ya &
garken zaki sun ƙunshi maza manya ɗaya zuwa uku masu dangantaka tare da mata da dama har zuwa talatin tare da yaya &
No error &
The only difference is the use of special Hausa characters. \\
\midrule
dong ɗan kasar koriya ne &
dung ɗan ƙasar koriya ne &
No error &
The only difference is the use of special Hausa characters. \\
\midrule
manyan jami’ai ne kawai suka samu damar shiga wurin shugaban kasar &
manyan jami'ai ne kawai suka samu damar shiga wurin shugaban ƙasar &
No error &
The only difference is the use of special Hausa characters. \\
\midrule
dalibai sun yi wasan kwallo &
ɗalibai sun yi wasan ƙwallo &
No error &
The only difference is the use of special Hausa characters. \\
\midrule
dalibai sun kai ziyara gidan masu tabin hankali &
ɗalibai sun kai ziyara gidan masu taɓin hankali &
No error &
The only difference is the use of special Hausa characters. \\
\midrule
dakin karatun yana dauke da dalibai kusan dubu daya &
ɗakin karatun yana ɗauke da ɗalibai kusan dubu ɗaya &
No error &
The only difference is the use of special Hausa characters. \\
\midrule
sai karfe tara na dare za'a sanar da sakamakon zaɓen &
sai ƙarfe tara na dare za a sanar da sakamakon zaɓen &
No error &
The only difference is the use of special Hausa characters. \\
\midrule
za'a yi mata aiki a ƙwaƙwalwa &
za a yi mata aiki a ƙwaƙwalwa &
Error in evaluation transcript &	
Grammatically, the correct form is `za a', not `za'a'. However, many people are using `za'a'. \\
\midrule
ana shan magani idan ba'a da lafiya &
ana shan magani idan ba a da lafiya &
Error in evaluation transcript &
Grammatically, the correct form is `ba a', not `a'ba'. However, many people are using `ba'a'. \\
\midrule
abuja na cikin nijeria &
abuja na cikin najeriya &
No error &
Both `Nijeriya' and `Naijeriya' are used, so it depends on the newspaper or individual. \\
\midrule
oguta karamin jiha ce a cikin nijeria &
oguta ƙaramin jiha ce a cikin najeriya &
No error &
The only difference is the use of special Hausa characters. \\
\midrule
sitika ɗin ƴana ɗa ƙyau &
sitika ɗin yana da kyau &
Error in evaluation transcript &
The only difference is the use of special Hausa characters. And there are wrong use of the special characters e.g ƴana The correct version is yana. \\
\midrule
ƴan shi'a sun yi tattaki jiya &
yan shi'a sun yi tattaki jiya &
No error &
The only difference is the use of special Hausa characters. \\
\midrule
macaroni abincin ƴan italiya ne &
makaroni abincin yan italiya ne &
No error &
The only difference is the use of special Hausa characters. \\
\midrule
kula da alaƙa mai ƙarfi da ƴan uwa &
kula da alaƙa mai ƙarfi da yan'uwa &
No error &
The only difference is the use of special Hausa characters. \\
\midrule
mallam aminu dan kasuwa ne à kasuwan kure &
malam aminu ɗan kasuwa ne a kasuwan kure &
Error in evaluation transcript &
Error in the evaluation transcript. There is no à in Hausa wrting sytle that we physically see and read. \\
\bottomrule
\end{longtable}

\begin{longtable}{p{0.25\textwidth} p{0.25\textwidth} p{0.15\textwidth} p{0.3\textwidth}}
\caption{Human evaluation of ASR model errors in Dholuo.} \\
\label{tab:luo_human_asr_eval} \\
\toprule
\textbf{Evaluation transcript} &
\textbf{Model output} &
\textbf{Evaluator assessment} &
\textbf{Evaluator comments} \\
\midrule
\endfirsthead
\caption[]{(continued) Human evaluation of ASR model errors in Dholuo.} \\
\toprule
\textbf{Evaluation Transcription} & 
\textbf{Model output} & 
\textbf{Evaluator assessment} & 
\textbf{Evaluator comment} \\
\midrule
\endhead

\midrule
\multicolumn{4}{r}{{Continued on next page}} \\
\midrule
\endfoot

\bottomrule
\endlastfoot

e piny kenya mano en ketho maduong' ahinya & e piny kenya mano en ketho maduong' ahinya & No error & Written Luo uses apostrophe at the final syllable as in the word maduong' but this does not result in a difference in meaning. \\
\midrule
tuwo mar sukari ema ne onego owadawano & tuo mar sukari ema ne onego owadwano & No error & Excellent, in Luo we either say `tuo' or `tuwo'. \\
\midrule
e wi mano tuwo mar corona ne oketho chenro mag somo e pinje & e wi mano tuo mar corona ne oketho chenro mag somo e pinje & No error & \\
\midrule
otho sa adek okinyi & otho saa adek okinyi & No error & No error, we either use `sa' or `saa'. \\
\midrule
neru maduong' osekendo & neru maduong’ osekendo & No error & Only difference in apostrophe used. \\
\midrule
seche moko ginyalo bedo jii ariyo ma penjo penjo & seche moko ginyalo bedo ji ariyo ma penjo penjo & No error & No error, it's either `ji' or `jii'. This a feature in Luo for monosyllabics.\\
\midrule
dhii uywe kund dhok & dhi uywe kund dhok & No error & See comments above. \\
\midrule
welo dhii e kanisa kawuono & welo dhi e kanisa kawuono & No error & See comments above. \\
\midrule
ngama kare ber & ng'ama kare ber & No error & Native speakers know this and would understand. \\
\midrule
unega kayiem nang’o & unega kayiem nang'o & No error & \\
\midrule
pesa jadoung ile & pesa jadoung’ ile & No error & Again optional final apostrophe. \\
\midrule
ng'ama nigi jadoung machiegni & ng'ama nigi jadoung’ machiegni & No error & \\
\midrule
en chieng' maduong & en chieng’ maduong’ & No error & \\
\midrule
antie gi othinyo mangeny & antie gi othinyo mang’eny & No error & \\
\midrule
we bedo gi gombo mangeny & we bedo gi gombo mang’eny & No error & \\
\midrule
mol mar trafik en timo nonro kata puonjruok kuom timbe mag joriembo kod mtokni e seche ma gisudo e kind kuonde ariyo to kod tudruoge magitimo e kindgi giwegi & mol mar trafik en timo nonro kata puonjruok kuom timbe mag joriembo kod mtokni e seche magisudo e kind kuonde ariyo to kod tudruoge magitimo e kindgi giwegi & No error & No error, `u' is an alternative for `wu'. \\
\midrule
onge wach achiel e piny duto ma lero tiend kaka mwandu molosi ichako luongi ni gima nyachon kembe moko mag solo osuru lero kuom mwandu mosteetieko higni maloyo 100 kaka gigo ma nyachon & onge wach achiel e piny duto malero tiend kaka mwandu molosi ichako luongi ni gima nyachon kembe moko mag solo osuru lero kuom mwandu mosteetieko higni maloyo 100 kaka gigo ma nyachon & No error & \\
\midrule
jarieko manyinge aristolet nowacho ni gik moko olos kod riwo achiel ariyo kata ang'wen mag gigi piny pii muya kod mach & jarieko manyinge aristotle nowacho ni gik moko olos kod riwo achiel ariyo kata ang'wen mag gigi piny pii muya kod mach & No error & See my comment on monosyllabics. \\
\midrule
atom moko nitiere kod nyukila ma ok ochung' motegno ma tiende ni gibarore ga ka otwomgi matin kata ka ok otwomgi chutho & atom moko ni tiyoore kod nyukila ma ok ochung’ motegno matiende ni gibarore ga ka otwomgi matin kata ka ok otwomgi chutho & No error & \\
\bottomrule
\end{longtable}

\begin{longtable}{p{0.25\textwidth} p{0.25\textwidth} p{0.15\textwidth} p{0.3\textwidth}}
\caption{Human evaluation of ASR model errors in Chichewa.} \\
\label{tab:luo_human_asr_eval} \\
\toprule
\textbf{Evaluation transcript} &
\textbf{Model output} &
\textbf{Evaluator assessment} &
\textbf{Evaluator comments} \\
\midrule
\endfirsthead
\caption[]{(continued) Human evaluation of ASR model errors in Chichewa.} \\
\toprule
\textbf{Evaluation Transcription} & 
\textbf{Model output} & 
\textbf{Evaluator assessment} & 
\textbf{Evaluator comment} \\
\midrule
\endhead

\midrule
\multicolumn{4}{r}{{Continued on next page}} \\
\midrule
\endfoot

\bottomrule
\endlastfoot

pa maulendo ena makampani ena akuluakulu ali ndi ndege zao koma pa maulendo ena makampani ang'onoang'ono amakhala ndi vuto & pamaulendo ena makampani ena akuluakulu ali ndi ndege zawo koma pamaulendo ena makampani ang'onoang'ono amakhala ndi vuto & Error & Model Output: grammar rules requires that pamaulendo written disjunctively as it is not a locative partical. Transcript: zao is gramatically wrong, should be written as zawo \\
\midrule
ana okulira kwaokha osakumana ndi anthu akhoza kutheka kuchitilidwa nkhanzza kapena kuzunzidwa asanasiyeidwe kapena kuthawa & ana okulira kwaokha osakumana ndi anthu akhoza kutheka kuchitilidwa nkhanzza kapena kuzuzidwa asanasiyeidwe kapena kuthawa & Error & Model Output: kuzuzidwa is a spelling error, it should be kuzunzidwa. \\
\midrule
ma blog amathanidzaniso ana asukulu kuphunzila kulemba ngakhalke kuti poyamba ophunzila amayamba ndi kulakwista galamala ndi zilembo za mawu kupeza kwa anthu omweolega kumathandizila kusintha izi & ma blogamothandizaso ana asukulu kuphunzila kulemba ngakhalke kuti poyamba ophunzila amayamba ndi kulakwitsa galamala ndi zilembo zamawu kupeza kwa anthu owenerenga kumathandizila kusintha izi & Error & Model Output: blogamothandizaso is unknown word made from combination of two or three words. This would confuse the reader. Amayambamba is a wrong spelling, it should be as in Transcription: amayamba. \\
\midrule
ngoziyi inachitikira mwamba m'mapiri atali ndipo akukhulupilira kuti zinachitika chifukwa cha adani achiwembu & ngoziyi inachitikira m'mwamba m'mapiri atali ndipo akukhulupilira kuti zinachitika chifukwa cha adani achiwembu & Error & Transcription: atali should be aatali as in model output. a chibwembu is normally written conjunctively and should be achibwembu as in Model Output. \\
\midrule
ku barcelona chiyankhulo chovomelezeka ndi catalan ndikli sipanishi theka la anthu akudziwa catalan ambiri amachimvetsa ndiipo pafupifupi onse amamva ndikudziwsa chi sipanishi & ku barcelona chiyankhulo chovomelezeka ndi catalan ndi chi sipanishi theka la anthu akudziwa catalan ambiri amachimvetsa ndiipo pafupifupi onse amamva ndikudziwsa chi sipanishi & Error & Transcription: chi should not be combined with chi but rather goes together with the name of the language to read: chisapnishi. Chi and sipanishi should be written conjunctively. Model Output: Chi and sipanishi should be written conjunctively. \\
\midrule
zolegeza zathawi zonse mu metro zimapangidwa muchilankhulo chachi kalatani basi koma zosiyanasiyana zimasulutidwa kudzera makina a kompyuta mu zilankhulo zosiyanasiyanansiya kuphatikizikachisapanishi chingerezi falansa arabic ndi japanese & zolegeza za nthawi zonse mu metro zimapangidwa muchilankhulo chachi kalatani basi koma zosiyanasiyana zimasulutidwa kudzera makina a kompyuta mu zilankhulo zosiyanasiyana kuphatikizika chisipanishi chingerezi falansa arabic ndi japanese & Error & Transcription: chi should not be combined with chi but rather goes together with the name of the language to read: chisapnishi. Kompyuta should be kompyuta. chisipanishi chingerezi falansa arabic ndi japanese should have commas in between. \\
\midrule
owona zangozi zokugwa madziidzi ku mpoto kwa mariana ati palibe zomve zidanowoneka pomwe analengezera a nation & owona zangozi zokugwa mwaziizizi kumpoto kwa mariana ati palibe zomve zidanowoneka pomwe analengezera a nation & Error & Transcription: madziidzi is an incomplete word which should read mwadziidzi. Model Output: Mwazizizi is gramatically incorrect. \\
\midrule
apia ndi likulu la samoa tauinyi ili pachilumba cha upolu ndipo pali chilengedwe chachilengedwe cha anthu ochepera pa 40000 & apia ndi likulu la samoa tauinyi ili pachilumba cha upolu ndipo pali chilengedwe chachilengedwe cha anthu ochepera pa 4000 & Error & Model Output: City of a pia should be Apia and the a should not be separated from pia. \\
\midrule
safari ndi mawu amene amatchulidwa kawirikawiri ndiipo amatanthauza ulendo wopamtunda wokaona nyama zokongola zakutchire za ku africa kawirikawiri ku savanna & safari ndi mawu amene amatchulidwa kawirikawiri ndiipo amatanthauza ulendo wopamtunda wokaona nyama zokongola zakutchire zaku africa kawirikawiri ku savanna & Error & Model Output: zaku should be written separately as za ku. \\
\midrule
kutenga sitima za m'madzi kunyamulira katundu ndi njira yoyenera zedi yonyamulira anthu wochuluka komanso katundu kuwoloka pa nyanja & kutenga sitima za m'madzi kunyamulira katundu ndi njira yoyenera zedi yonyamulira anthu ochuluka komanso katundu kooloka panyanja & Error & Model Output: kooloka is wrong spelling of kuwoloka \\
\midrule
mkulu wophunzitsa pa sukulu ya ukachenjede ya dundee university a pulofesa pamela ferguson adati atolankhani akuoneka kuti akuyenda mu chiwopsezo akamasindikiza zithunzi ndi zina zotero za oganizilidwa kupalamula milandu & mgalu wophunzitsa pasukulu yaukachenjede ya dud university a pulofesa pamella fegason adati atolankhani akuoneka kuti akuyenda mu chiopsezo akamasindikiza zithuzi ndi zina zotero za oganizilidwa kupalamula milandu & Error & Model Output: mgulu is wrong spelling of mkulu. Yaukachenjede should be written as ya ukachenjede \\
\midrule
thandizo loiyikidwa m'maphunziro apa kompyuta ndipo akuyenera kufunsa kupanga zina zakefotokozera ndondomeko zomwe zikanakhala zovuta kwa ophunzira & thandizo loiyikidwa m'maphunziro apakompyuta ndipo akuyenera kufunsa kupanga zina zake ndikufotokozera ndunudmiko zomwe zikanakhala zovuta kwa ophunzira & Error & Model Output: ndunudmiko is wrong spelling of ndondomeko and kompyuta should read kompyuta. \\
\midrule
bomba la fission limagwira ntchito pamene limafuna mphamvu kuti liyike pamodzi ma nucleus wochulukana ndi ma proton ambiri ndi ma neutron & bomba la fission limagwira ntchito pamene limafuna mphamvu kuti liyike pamodzi ma nucleus wochulukana ndi ma proton ambiri ndi ma neutron & No error & \\
\midrule
ma ion ndima proton a hydrogen amathotholedwa popeza hydrogen amakhala ndi proton imodzi ndi electron imodzi & ma ion ndi ma proton a hydrogen amathotholedwa popeza hydrogen amakhala ndi proton imodzi ndi electron imodzi & Error & Model Output: amathotoleledwa is wrong spelling of amathotholedwa \\
\midrule
wakhalapo akulepherera kumwa mankhwala ofunikira pochiza ululu omwe akumva chifukwa cha matenda alowedwa pa masewera & wakhalapo akulepherera kumwa mankhwala ofunikira pochiza ululu omwe akumva chifukwa cha matenda olestedwa pamasewera & No error & Transcription: Woyima should be oyima. Model Output: Kuwumikizana is wrong spelling of kulumikizana. Maro is wrong spelling of Malo. \\
\midrule
zilumba zambiri zing'onoang'ono ndi mayiko woyima powapita kulumikizana ndi dziko la france ndi ziko la arabic ndi japanese & zilumba zambiri zing'onoang'ono ndi mayiko woyima powapita kulumikizana ndi dziko la france ndi ziko la arabic ndi japanese & Error & Transcription: Woyima should be oyima. Model Output: Kuwumikizana is wrong spelling of kulumikizana. Maro is wrong spelling of Malo. \\
\midrule
kuwonetsera kwa nyumba zomwe zimapangidwa mawonedweko a hong kong skyline akutchulidwa victoria harpur bar tatchi yowala kwambiri m'madera oyandikira doko chikwangwanzi akamapereka zikwangwanzi m'dera lozungulira zonyamula anthu omwe amafika mochuluka & kuwonetsera kwa nyumba zomwe zimapangidwa mawonedweko a hongkong skyline akutchulidwa victoria harpur bar tatchi yowala kwambiri m'madera oyandikira doko chikwangwanzi akamapereka zikwangwanzi mdera lozungulira zonyamula anthu omwe amafika mochuluka & Error & Model Output: victoria harpur is misspelling of victoria harbour. \\
\midrule
kujambula makamera amachita kusiyana atakhala pa kompyuta pakumasulira kwa mwachangu amene ang'ono chinthu chinasache kupeza kwa linako kapena wotsekereza monga ndondomeko yomwe ingathe kupangidwa ndi anthu ochepa & kujambula makamera amachita kusiyana atakhala pa kompyuta pakumasulira kwa mwachangu amene ang'ono chinthu chinasache kupeza kwa linako kapena wotsekereza monga ndondomeko yomwe ingathe kupangidwa ndi anthu ochepa & Error & Transcription: Milisecond should be transliterated to Milisekondi. Model Output: miliceand is a wrong spelling of millisecond which is normally transliterated as milisecond. \\
\midrule
pamene mafumu ndi akuluakulu mabanja ndi zochitika mufunikanso kuti mufikeko nsanga ngati kuli msanga apafupi ndi nyumba & pamene mafumu ndi akuluakulu mabanja ndi zochitika mufunikanso kuti mufikeko nsanga ngati kuli msanga apafupi ndi nyumba & Error & Model Output: mufikeko is wrong spelling of mufikeko. \\
\midrule
pakuti mu nthawi yao kuonjeza kuwala kwa sikunali wuto monga anali pa makolo m'pang'ono amafunika kuwala koopsa kufikila kusiyana ndi omwe amaganidwa makono ano & pakuti mu nthawi yao kuonjeza kuwala kwa sikunali wuto monga anali pa makolo m'pang'ono amafunika kuwala koopsa kufikila kusiyana ndi omwe amaganidwa makono ano & Error & Model Output: panu is wrong spelling of pano. Campus should be transliterated to kampasi or just describe what a campus is. \\
\bottomrule
\end{longtable}

\renewcommand{\thesection}{Appendix \Alph{section}}
\renewcommand{\thefigure}{F.\arabic{figure}}
\setcounter{figure}{0} 
\renewcommand{\thetable}{F.\arabic{table}}
\setcounter{table}{0} 

\clearpage 
\section{Wav2Vec-BERT 2.0 Hyperparameters}
~\label{sec:appendix_mms_asr_results}

\begin{table}[ht]
\centering
\begin{tabular}{lc}
\hline
\textbf{Hyperparameter} & \textbf{Value} \\
\hline
Learning rate & 3e-05 \\
Warmup ratio & 0.1 \\
Evaluation steps & 1000 \\
Early stopping patience & 5 \\
Add adapter & True \\
Mask time probability & 0 \\
Attention dropout & 0.05 \\
Feature projection dropout & 0.05 \\
Hidden layer dropout & 0.05 \\
CTC zero infinity & True \\
\hline
\end{tabular}
\caption{Common Wav2Vec-BERT 2.0 hyperparameters.}
\label{tab:common_w2v_hyperparameters}
\end{table}

\begin{table}[ht]
\centering
\begin{tabular}{lcc}
\hline
\textbf{Real:Synth Ratio} & \textbf{(Maximum) Number of epochs} & \textbf{(Total) Batch size} \\
\hline
100h:0     & 250 & 320 \\
250h:0     & 100 & 320 \\
500h:0     & 50  & 320 \\
100h:400h  & 50  & 320 \\
250h:250h  & 50  & 320 \\
579h:450h  & 24  & 320 \\
579h:993h  & 16  & 320 \\
\hline
\end{tabular}
\caption{Hausa Wav2Vec-BERT 2.0 Hyperparameters. We keep epoch-hours constant, that is the number of epochs multiplied by the total duration of the training dataset in hours.}
\label{tab:hausa_w2v_hyperparameters}
\end{table}

\begin{table}[ht]
\centering
\begin{tabular}{cc}
\hline
\textbf{(Maximum) Number of steps} & \textbf{(Total) Batch size} \\
\hline
100000 & 64 \\
\hline
\end{tabular}
\caption{Dholuo and Chichewa Wav2Vec-BERT 2.0 hyperparameters.}
\label{tab:luo_chichewa_w2v_hyperparameters}
\end{table}

\end{document}